\documentclass[letterpaper]{article} 
\usepackage{aaai24}  
\usepackage{times}  
\usepackage{helvet}  
\usepackage{courier}  
\usepackage[hyphens]{url}  
\usepackage{graphicx} 
\usepackage{natbib}  
\usepackage{caption} 
\usepackage{algorithm}
\usepackage{algorithmic}

\usepackage{newfloat}
\usepackage{listings}
\DeclareCaptionStyle{ruled}{labelfont=normalfont,labelsep=colon,strut=off} 
\floatstyle{ruled}
\newfloat{listing}{tb}{lst}{}
\floatname{listing}{Listing}
\usepackage{times}  
\usepackage{helvet}  
\usepackage{courier}  
\usepackage[hyphens]{url}  
\usepackage{graphicx} 
\usepackage{natbib}  
\usepackage{caption} 
\usepackage{subcaption}
\usepackage{multirow}
\usepackage{tikz}
\usepackage{algorithm}
\usepackage{algorithmic}
\usepackage{lipsum}

\usepackage{newfloat}
\usepackage{listings}
\DeclareCaptionStyle{ruled}{labelfont=normalfont,labelsep=colon,strut=off} 
\floatstyle{ruled}
\newfloat{listing}{tb}{lst}{}
\floatname{listing}{Listing}
\usepackage{graphics}
\usepackage{amsmath}
\usepackage[inkscapeformat=png]{svg}

\usepackage[capitalize]{cleveref}
\crefname{section}{Sec.}{Secs.}
\crefname{subsection}{Sec.}{Secs.}
\crefname{table}{Tab.}{Tabs.}
\crefname{Figure}{Fig.}{Figs.}
\crefname{Equation}{Eq.}{Eqs.}

\newcommand{\chesstable}{\textsc{Chess Table}~}
\newcommand{\colorfountain}{\textsc{Color Fountain}~}
\newcommand{\stove}{\textsc{Stove}~}
\newcommand{\shoerack}{\textsc{Shoe Rack}~}

\newcommand{\horns}{\textsc{Horns}~}
\newcommand{\trex}{\textsc{T-Rex}~}
\newcommand{\flower}{\textsc{Flower}~}
\newcommand{\orchids}{\textsc{Orchids}~}
\newcommand{\fortress}{\textsc{Fortress}~}
\newcommand{\fern}{\textsc{Fern}~}

\begin{document}

\title{GSN: Generalisable Segmentation in Neural Radiance Field}
\author {
    Vinayak Gupta\textsuperscript{\dag\rm 1},
    Rahul Goel\textsuperscript{\rm 2},
    Sirikonda Dhawal\textsuperscript{\rm 2},
    P. J. Narayanan\textsuperscript{\rm 2}
}
\affiliations {
    \textsuperscript{\rm 1}Indian Institute of Technology, Madras \\\ \ \ \ 
    \textsuperscript{\rm 2}International Institute of Information Technology, Hyderabad\\
    vinayakguptapokal@gmail.com,
    \{rahul.goel,dhawal.sirikonda\}@research.iiit.ac.in,
    pjn@iiit.ac.in
}

\maketitle
\let\thefootnote\relax\footnote{$^\dag$ Work done during internship at IIIT Hyderabad.}

\begin{abstract}
Traditional Radiance Field (RF) representations capture details of a specific scene and must be trained afresh on each scene. Semantic feature fields have been added to RFs to facilitate several segmentation tasks. Generalised RF representations learn the {\em principles} of view interpolation. A generalised RF can render new views of an {\em unknown and untrained} scene, given a few views. We present a way to distil feature fields into the generalised GNT representation. Our GSN representation generates new views of unseen scenes on the fly along with consistent, per-pixel semantic features. This enables multi-view segmentation of arbitrary new scenes. We show different semantic features being distilled into generalised RFs. Our multi-view segmentation results are on par with methods that use traditional RFs. GSN closes the gap between standard and generalisable RF methods significantly. Project Page: \url{https://vinayak-vg.github.io/GSN/}

\end{abstract}
\section{Introduction}
{
    \label{sec:introduction}
    Capturing, digitising, and authoring detailed structures of real-world scene settings is tedious. The digitised real-world scenes open many applications in graphics and augmented reality. Efforts to capture real-world scenes included special hardware setup \cite{PJN:ICCV98}, structure from motion (SfM) from unstructured collections \cite{phototourism, romein1day}, and recently to Neural Radiance Fields (NeRF) \cite{nerf}. NeRF generates photo-realistic novel views using a neural representation learned from a set of discrete images of a scene. Follow-up efforts enhanced performance, reduced memory, etc.\ \cite{DFF, stylizednerf_cvpr, instantngp, tensorf}. Grid-based representations led to faster learning \cite{plenoxels,dvgo,tensorf}. These use scene-specific representations that do not generalise to unseen scenes.
    
    PixelNeRF \cite{pixelnerf}, IBRNet \cite{ibrnet}, MVSNeRF \cite{MVSNeRF} and GNT \cite{varma2022attention} generalise RFs to unknown scenes by formulating novel view synthesis as multi-view interpolation. PixelNeRF introduced a scene-class prior and used a CNN to render novel views. IBRNet used an MLP and a ray transformer to estimate radiance and volume density, collecting information on the fly from multiple source views. MVSNeRF builds on the work of MVSNet \cite{yao2018mvsnet} to create a neural encoding volume based on the nearest source views. GNT improves the IBRNet formulation to leverage epipolar constraints using attention-based view transformers and ray transformers.

    Radiance Fields are trained using a few dozen views of a scene and ``memorise'' a {\em specific scene}. New views are generated from it using a volumetric rendering procedure. Generalised RF methods, on the other hand, learn the {\em principles} of view generation using epipolar constraints and interpolation of proximate views. They are trained using views from multiple scenes. Novel views of an {\em unknown scene} for any camera pose are generated by directly interpolating a few views. The distinction between learning a scene and learning a view-based rendering algorithm is significant.

    Users are often interested in segmenting parts of the captured 3D scene and manipulating them, besides new view generation. Initial segmentation efforts like N3F \cite{N3F} and DFF \cite{DFF} distilled a semantic field alongside a radiance field. More recent ISRF \cite{isrf} leveraged proximity in 3D space and the distilled semantic space for interactive segmentation to extract good multi-view segmentation masks.
    
    Can we combine semantics into the generalised Radiance Fields to facilitate semantic rendering along with view generation of unseen scenes?
    In this paper, we extend the GNT formulation to include different semantic features efficiently. This makes the generation of consistent semantic features possible, along with generating new views of an unknown scene on the fly. Our generalisable representation can take advantage of semantic features like DINO \cite{dino}, CLIP \cite{clip}, and SAM \cite{sam}, to facilitate more consistent segmentation of the scene in new views. The key points about our method are
:
    \begin{itemize}
        \item Integrate semantic features into a generalised radiance field representation deeply, facilitating the rendering of per-pixel semantic features for each novel view. This is done without retraining or fine-tuning to new scenes.
        \item Facilitate object segmentation, part segmentation, and other operations simultaneously for new views, exploiting the per-pixel semantic features.
        \item Provide segmentation quality similar to the best scene-specific segmentation methods for radiance fields.
        \item Allow for distilling multiple semantic fields in the generalised setting. This results in finer and cleaner semantic features that perform better than the original.
    \end{itemize}



}

\section{Related Work}
{
    \label{sec:related_work}

    Since NeRF \cite{nerf} has come out, much research has been surrounding it. For a good overview of the study around radiance fields, we recommend the readers to refer to these excellent surveys \cite{nerf_survey, nerf_survey2}.

    \subsection{Neural Radiance Fields}
    {
        \label{sub_sec:related_work_nerf}
        Given a point $(x, y, z)$ in 3D and a viewing direction specified using polar angles $(\theta, \phi)$, a radiance field $\mathcal{F}$ maps these parameters to an RGB color(radiance) value: $\mathcal{F}(x, y, z, \theta, \phi): \mathcal{R}^3 \times \mathcal{S}^2 \rightarrow \mathcal{R}^3$.
    
        With assistance from a density field, these values can be accumulated along a ray shot through a pixel using the volumetric rendering equation \cref{eq:volume_rendering}.
               \begin{align}
                \label{eq:volume_rendering}
                \small
                \hat{C}(r) = \left( \textstyle\sum_{i=i}^{K} T_i \alpha_i c_i \right) \
                \qquad \textrm{where} \notag\\
                \alpha_i = 1 - e^{-\sigma_i \delta_i} 
                \qquad \textrm{and} \qquad
                T_i = \textstyle\prod_{j=1}^{i-1} (1 - \alpha_j).
                \end{align}
    
        In  \cref{eq:volume_rendering}, for a sampled point $i$ along a ray, $\delta_i$ is the inter-sample distance, $T_i$ is the accumulated transmittance, and $c_i$ is the directional colour for the point. For more details about this formulation, please refer to NeRF \cite{nerf}.
    
        NeRF models the density field and the radiance field using an MLP. Since the advent of NeRFs, there have been various works to improve NeRFs to handle reflections better \cite{verbin2022refnerf}, to handle sparse input views \cite{Niemeyer2021Regnerf, xu2022sinnerf}, improve quality in the case of unconstrained images \cite{martinbrualla2020nerfw, chen2022hallucinated}, to model large scale scenes \cite{tancik2022block, Turki_2022_CVPR}, in controllability for editing\cite{wang2022clip, yuan2022nerf} and to deal with dynamic scenes \cite{park2021hypernerf, kplanes}.
        
        Several efforts have attempted to improve efficiency and rendering speed of radiance fields using dense grids \cite{dvgo}, hash grids \cite{instantngp}, decomposed grids \cite{tensorf}, and gaussians \cite{gsplat}. However, all these methods perform learning/optimisation that is scene-specific and cannot generalise to new scenes.

    }

    \subsection{Generalised Neural Radiance Fields}
    {
        \label{sub_sec:related_work_generalized}
        A fundamental shortcoming of most radiance field methods is that they need to be trained separately for each scene, i.e., they are scene-specific. Multiple attempts have been made to generalize them to wild scenes without training. MVSNeRF \cite{MVSNeRF}, a deep neural network to reconstruct radiance fields by using constraints from multi-view stereo. Their approach generalises across scenes by constructing a neural encoding volume using a cost volume built on the three nearest source views and uses an MLP to regress the corresponding density and colour per point on the ray. IBRNet \cite{ibrnet} uses an MLP and a ray transformer that estimates the volumetric density and radiance for a given 3D point from a view direction. Instead of learning the scene structure, the neural networks learn how the source views of the scene can be interpolated to produce a new view. This allows them to generalise to arbitrary scenes. GNT \cite{varma2022attention} improves IBRNet \cite{ibrnet} using a series of view transformers and ray transformers. They completely substitute the volumetric rendering equation with a ray transformer and show that it acts as a better feature aggregator and produces superior results for the task of novel view synthesis.
    }

    \subsection{Semantics in Radiance Fields}
    {
        \label{sub_sec:related_work_semantics}
        Multi-view segmentation is a basic problem in computer vision. It could be a requirement for several downstream tasks \cite{neus}. Neural radiance fields are ideal for this task since they provide good-quality 3D reconstruction. Since we can render images from new views, it also enables segmentation from unseen views. Multiple efforts have successfully incorporated semantics into radiance fields. N3F \cite{N3F} showed incorporation of DINO \cite{dino} features into NeRFs, which can be used to perform segmentation. Concurrently, DFF \cite{DFF} distilled DINO features as well as LSeg \cite{lseg} features into a radiance field and portrayed similar results. LERF \cite{lerf} brings CLIP features into RFs to enable language queries similar to DFF. NVOS \cite{nvos} uses graph-cut to segment 3D neural volumetric representations. ISRF \cite{isrf} also proposes distilling DINO features into a voxelised radiance field. They show state-of-the-art segmentation by leveraging the spatial growth of a 3D mask. However, these efforts suffer from the same issue: their representation and semantic distillation are scene-specific. The model needs to be retrained and re-distilled for a each new scene.
    }

    Contrary to previous distillation works, we propose a method to distil features in a generalisable radiance field. This allows us to generate semantic feature images alongside RGB images for a new scene from any viewpoint. These semantic features can be used for several downstream tasks. In particular, we demonstrate their capability on a segmentation task. We also show that the features predicted by our generalised architecture are finer and cleaner than the original features of the images due to the flow of information from other nearby views.
}
\begin{figure*}[!ht]
    \centering
    \centering
    \includegraphics[width=\textwidth]{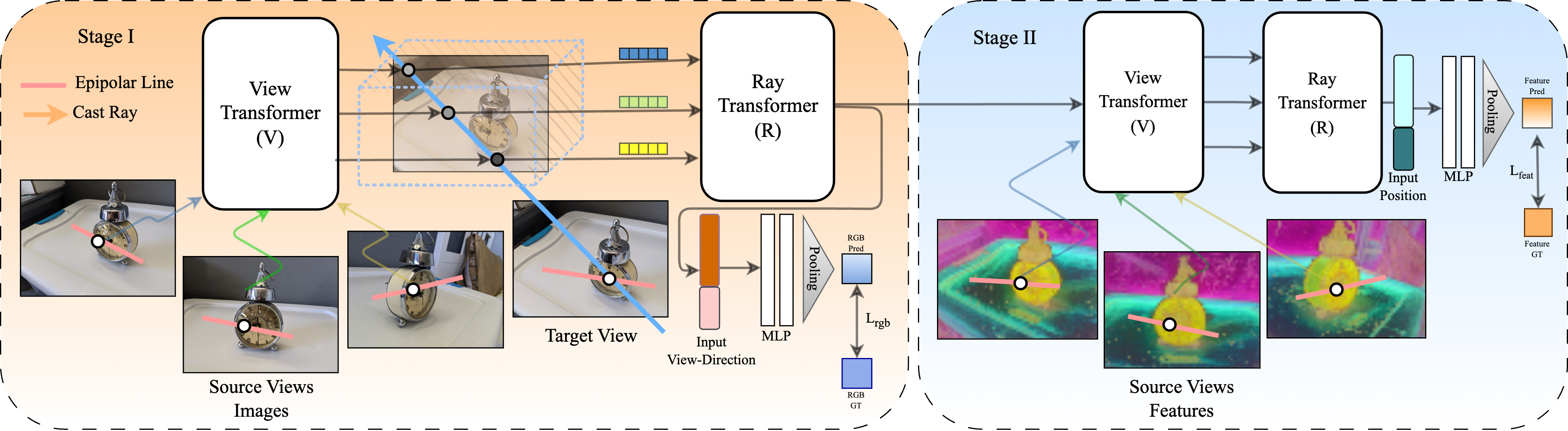}
    \label{modelarch}
    \caption{\label{fig:modelarch}{\em Overview:} Stage I: 1) We aggregate the features from the source views in View Transformer constrained by the epipolar geometry 2) The point aggregated features are passed on to the ray transformer along with input positions to aggregate the information along the ray. 3) The ray aggregated features and input view direction are passed onto an MLP and pooled to obtain pixel-wise colour. Stage II: 4) The view-independent features from the ray transformer are passed on to the stage-II block and aggregated by the view transformer and the ray transformer using the source view features extracted from the image using a pre-trained model like DINO. 5) The features out of the ray transformer are concatenated with input positions and pooled to predict pixel-wise features of the corresponding target-view pixel.
}
\end{figure*}

\begin{figure*}[!ht]
    \centering
    
    \begin{minipage}{0.24\linewidth}
        \centering
        \chesstable
    \end{minipage}
    \begin{minipage}{0.24\linewidth}
        \centering
        \colorfountain
    \end{minipage}
    \begin{minipage}{0.24\linewidth}
        \centering
        \stove
    \end{minipage}
    \begin{minipage}{0.24\linewidth}
        \centering
        \shoerack
    \end{minipage}
    
    \rotatebox[origin=c]{90}{Reference}
    \begin{minipage}{0.24\linewidth}
    \centering
    \frame{\includegraphics[width=\textwidth]{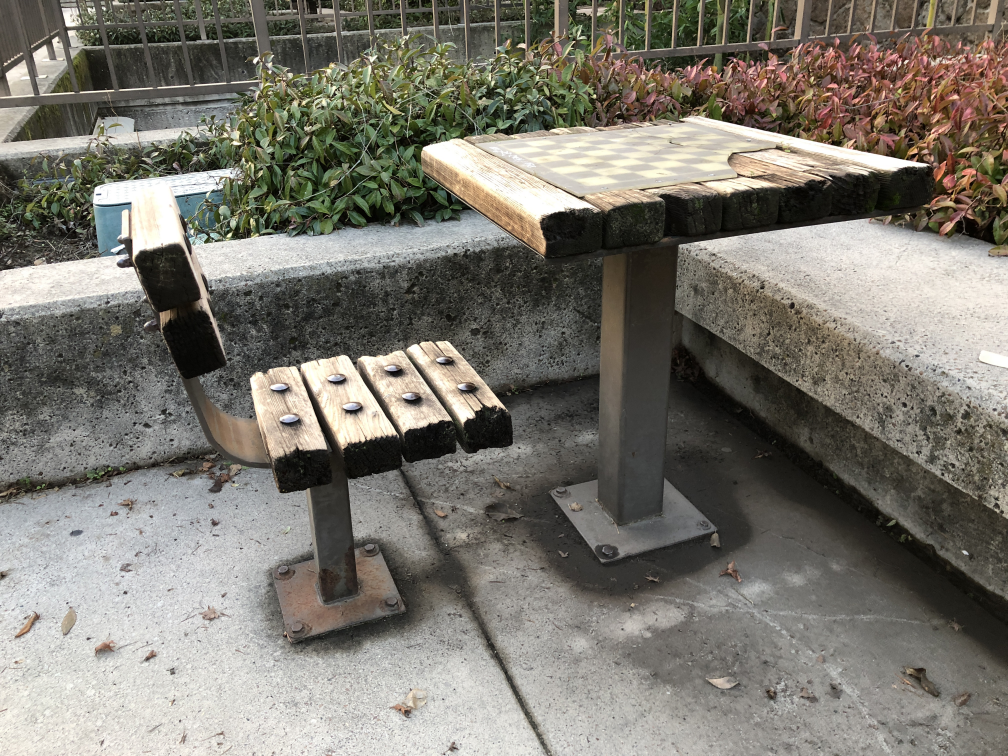}}
    \end{minipage}
    \begin{minipage}{0.24\linewidth}
        \centering
         \frame{\includegraphics[width=\textwidth]{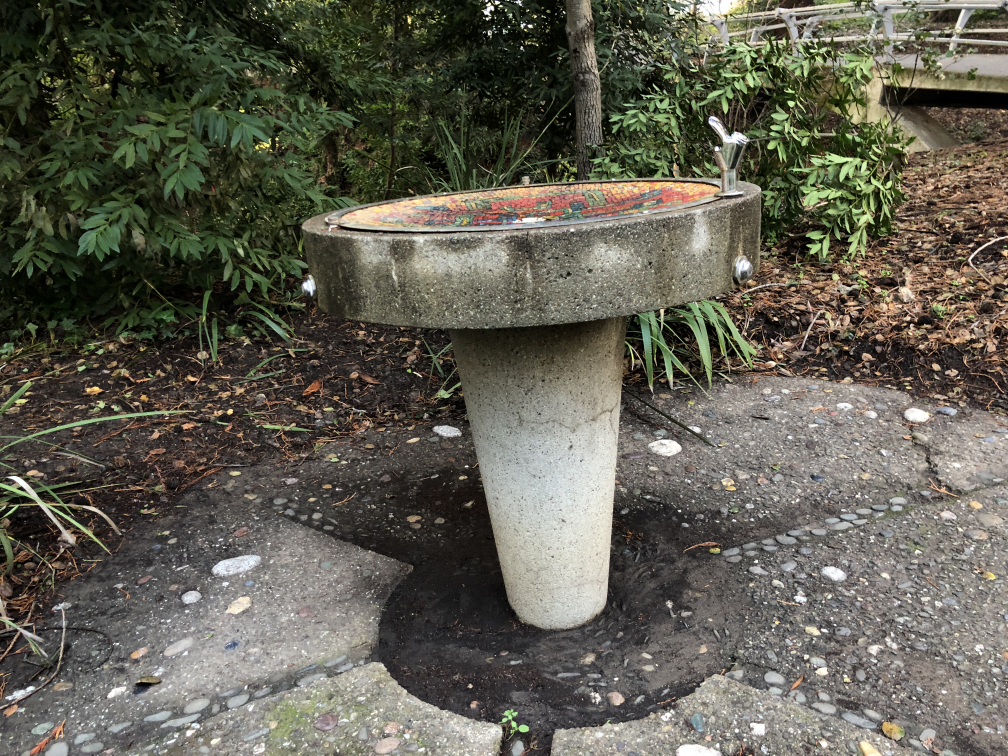}}
    \end{minipage}
    \begin{minipage}{0.24\linewidth}
        \centering
         \frame{\includegraphics[width=\textwidth]{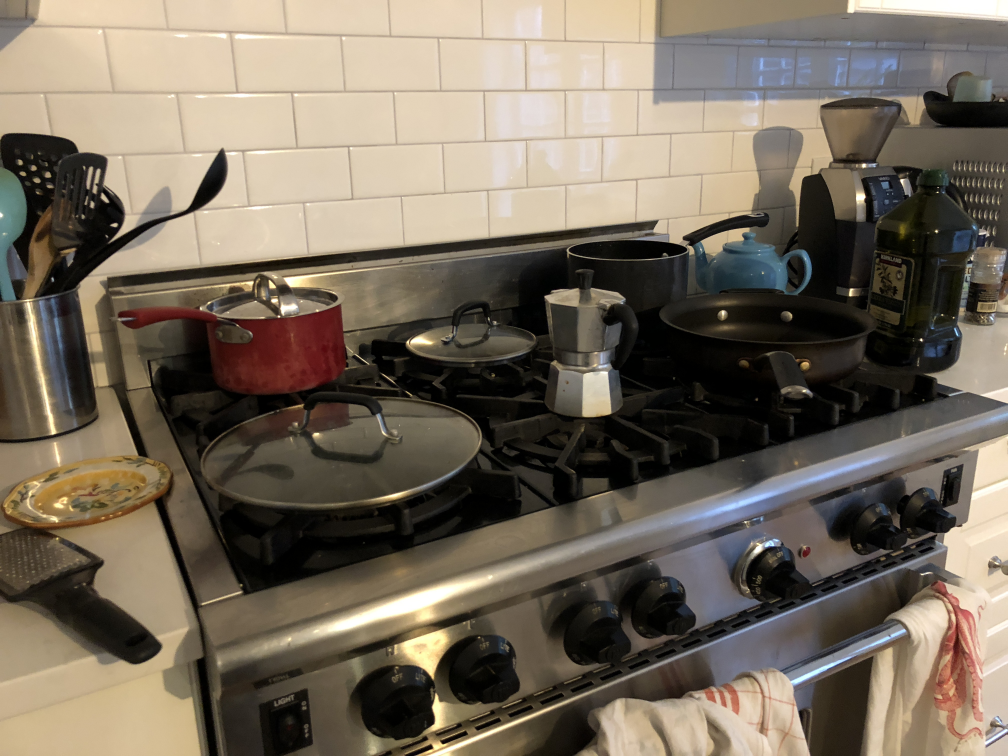}}
    \end{minipage}
    \begin{minipage}{0.24\linewidth}
        \centering
         \frame{\includegraphics[width=\textwidth]{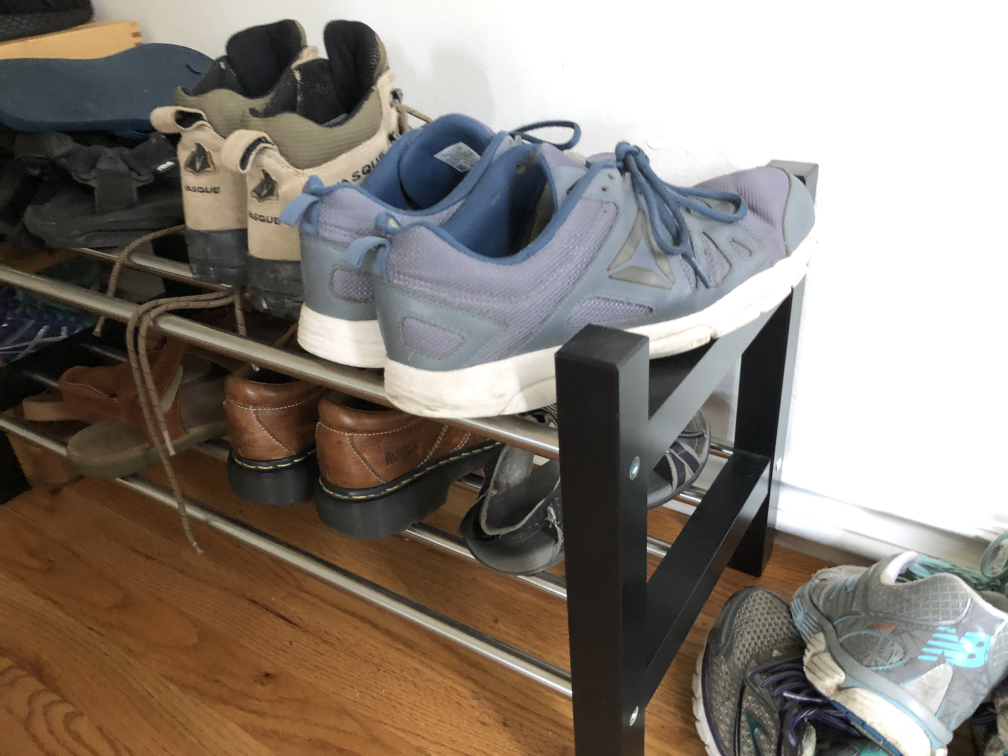}}
    \end{minipage}
    
    \rotatebox[origin=c]{90}{N3F/DFF}
    \begin{minipage}{0.24\linewidth}
        \centering
        \frame{\includegraphics[width=\textwidth]{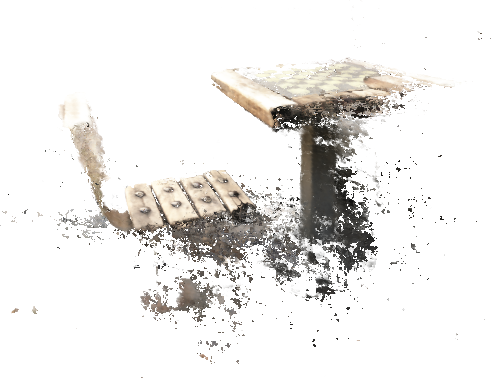}}
    \end{minipage}
    \begin{minipage}{0.24\linewidth}
        \centering
         \frame{\includegraphics[width=\textwidth]{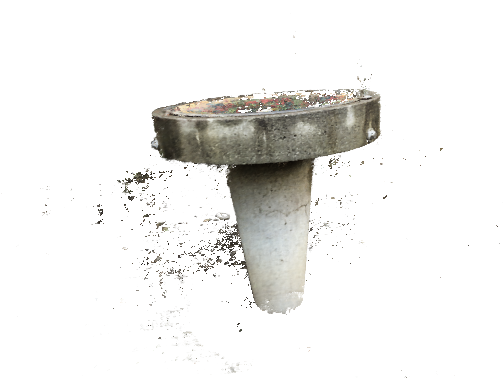}}
    \end{minipage}
    \begin{minipage}{0.24\linewidth}
        \centering
         \frame{\includegraphics[width=\textwidth]{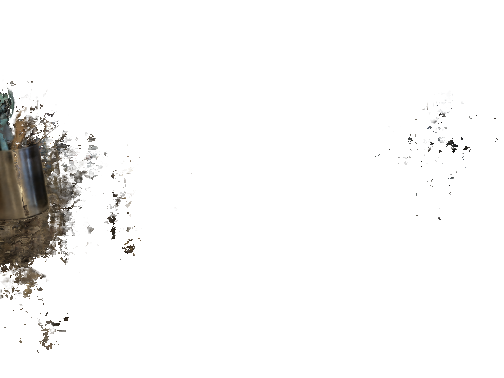}}
    \end{minipage}
    \begin{minipage}{0.24\linewidth}
        \centering
         \frame{\includegraphics[width=\textwidth]{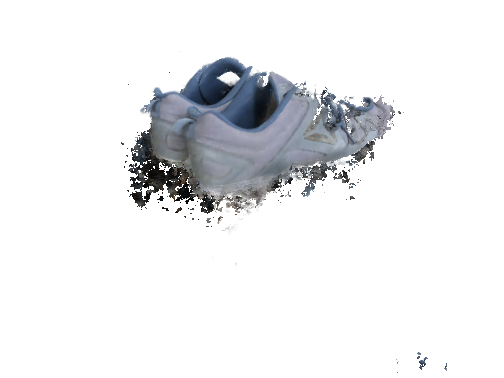}}
    \end{minipage}
    
    \rotatebox[origin=c]{90}{ISRF}
    \begin{minipage}{0.24\linewidth}
        \centering
        \frame{\includegraphics[width=\textwidth]{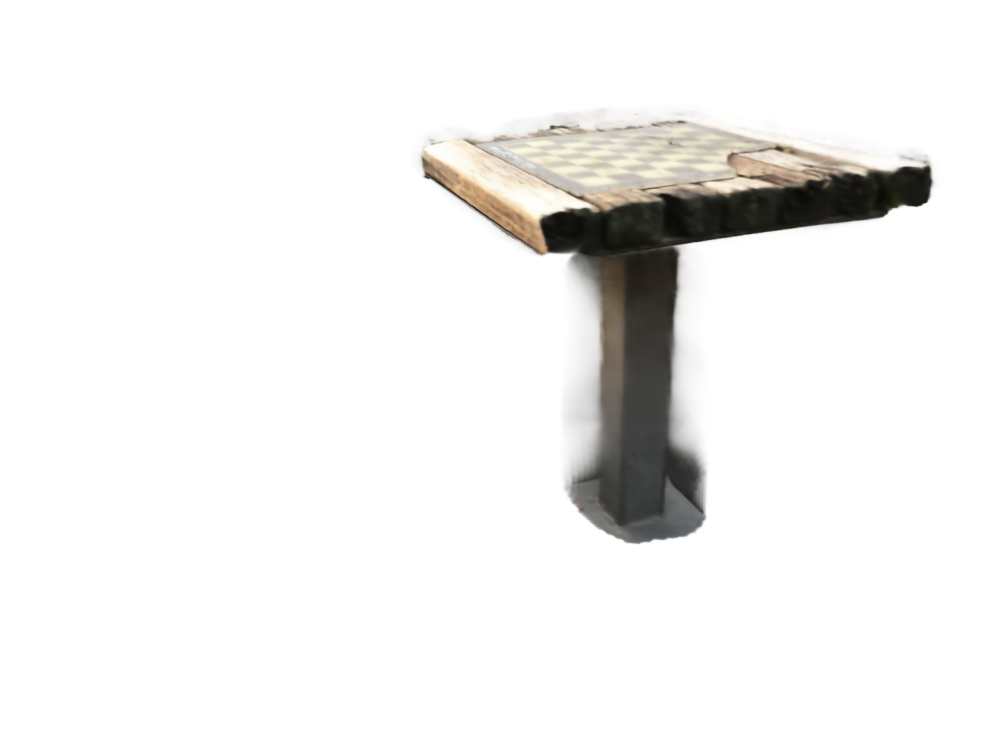}}
    \end{minipage}
    \begin{minipage}{0.24\linewidth}
        \centering
         \frame{\includegraphics[width=\textwidth]{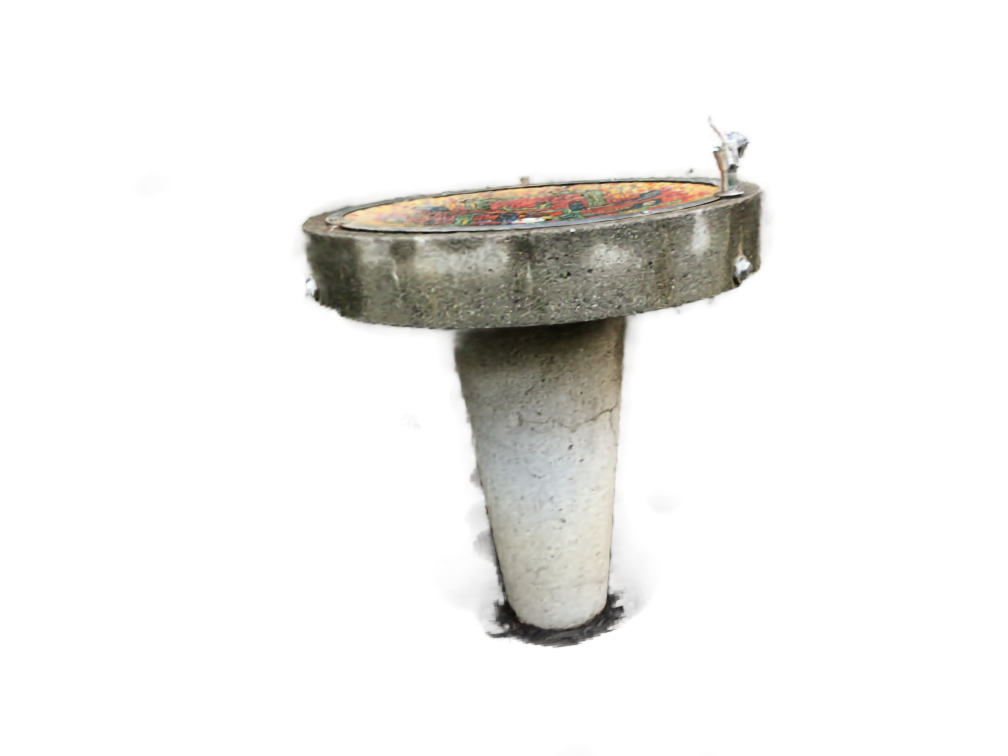}}
    \end{minipage}
    \begin{minipage}{0.24\linewidth}
        \centering
         \frame{\includegraphics[width=\textwidth]{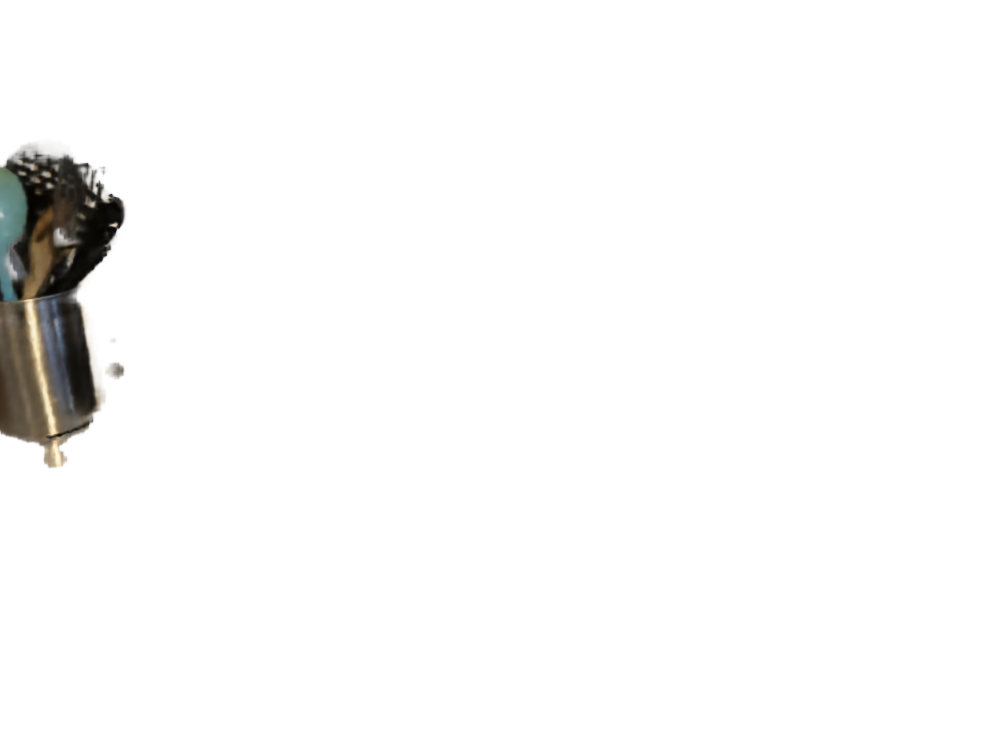}}
    \end{minipage}
    \begin{minipage}{0.24\linewidth}
        \centering
         \frame{\includegraphics[width=\textwidth]{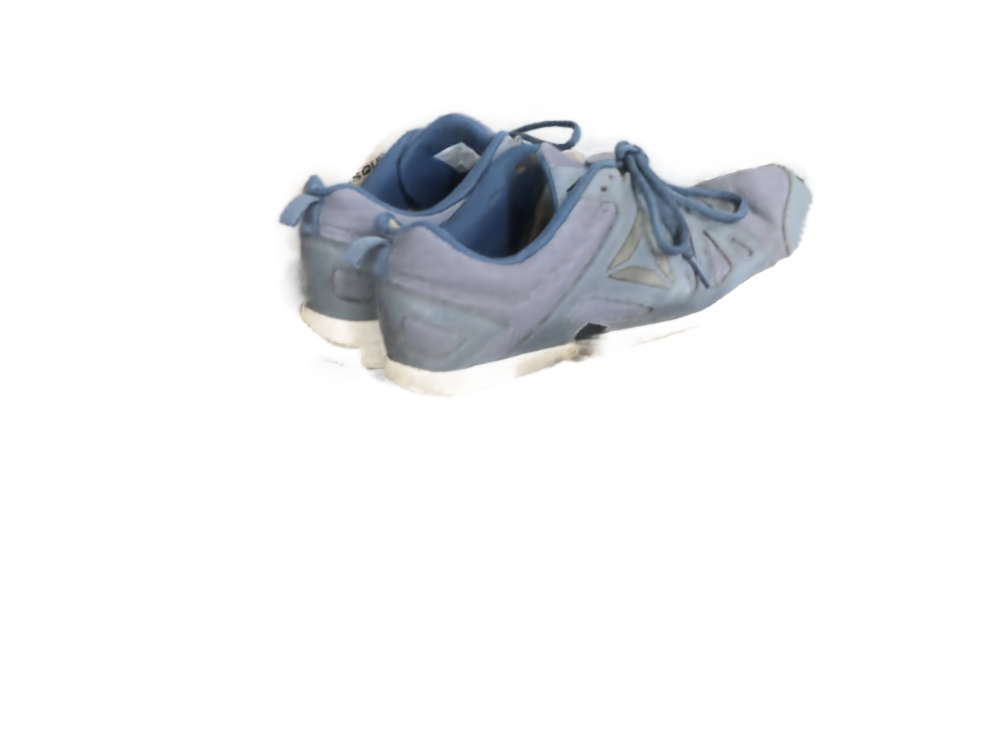}}
    \end{minipage}
    
    \rotatebox[origin=c]{90}{Ours}
    \begin{minipage}{0.24\linewidth}
        \centering
        \begin{tikzpicture}
            \node[anchor=south west,inner sep=0] (image) at (0,0) {\frame{\includegraphics[width=\textwidth]{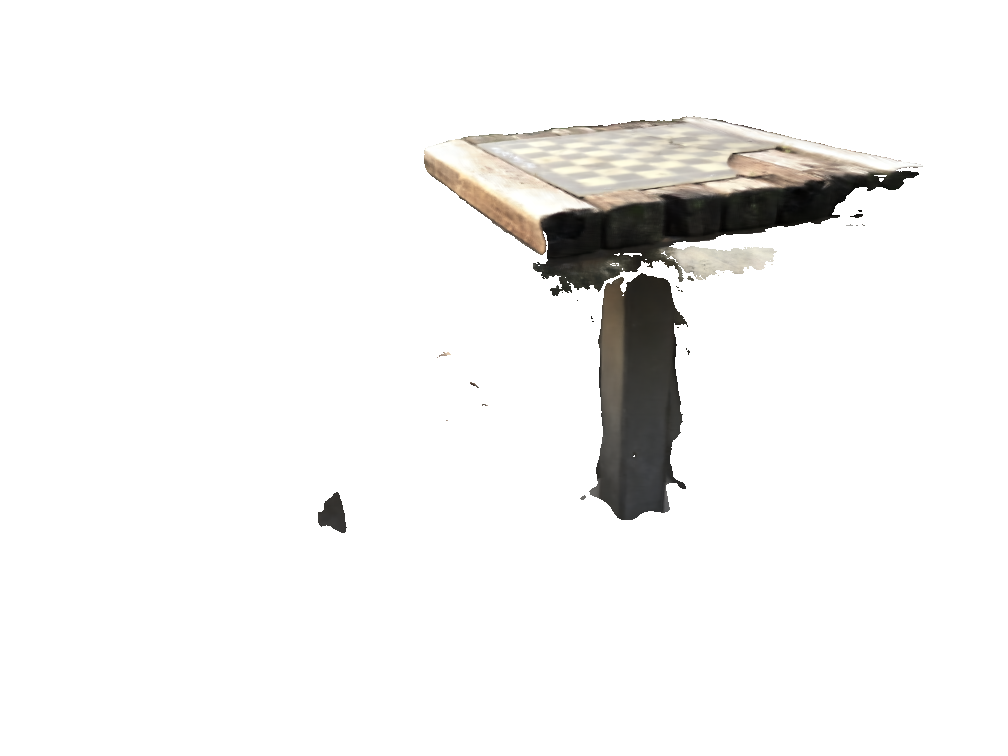}}};
            \begin{scope}[x={(image.south east)},y={(image.north west)}]
                \draw[red,thick] (0.5,0.6) rectangle (0.9, 0.75);
            \end{scope}
        \end{tikzpicture}
    \end{minipage}
    \begin{minipage}{0.24\linewidth}
        \centering
        \begin{tikzpicture}
            \node[anchor=south west,inner sep=0] (image) at (0,0) {\frame{\includegraphics[width=\textwidth]{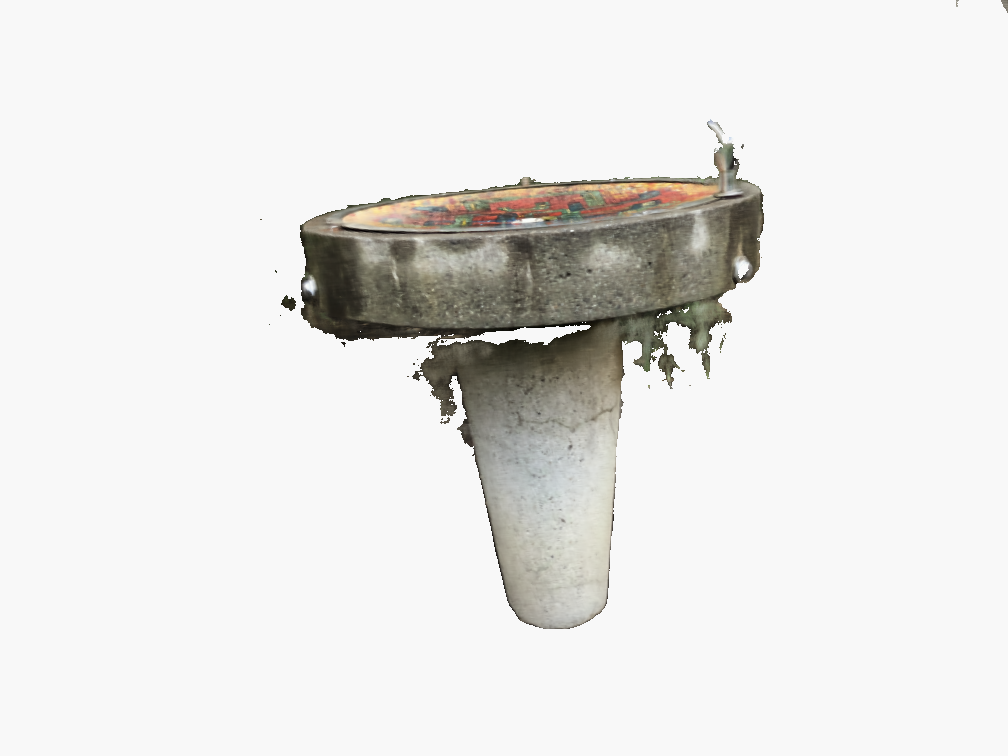}}};
            \begin{scope}[x={(image.south east)},y={(image.north west)}]
                \draw[red,thick] (0.35,0.5) rectangle (0.75, 0.6);
            \end{scope}
        \end{tikzpicture}
    \end{minipage}
    \begin{minipage}{0.24\linewidth}
        \centering
        \begin{tikzpicture}
            \node[anchor=south west,inner sep=0] (image) at (0,0) {\frame{\includegraphics[width=\textwidth]{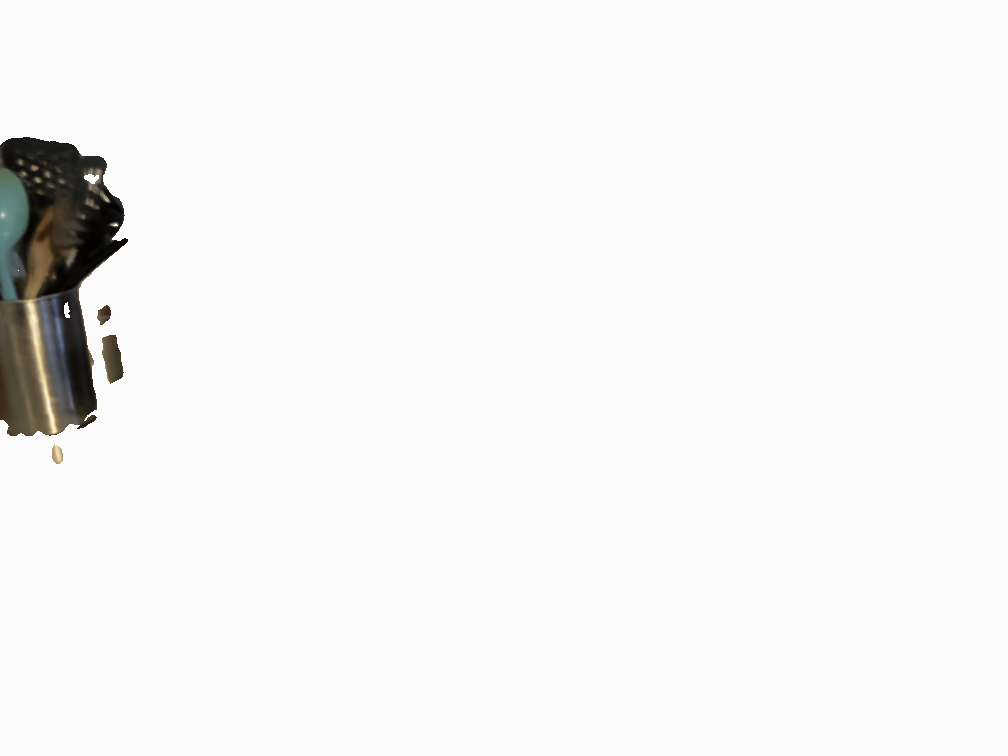}}};
            \begin{scope}[x={(image.south east)},y={(image.north west)}]
                \draw[green,thick] (0.02,0.68) rectangle (0.15, 0.85);
            \end{scope}
        \end{tikzpicture}
    \end{minipage}
    \begin{minipage}{0.24\linewidth}
        \centering
        \begin{tikzpicture}
            \node[anchor=south west,inner sep=0] (image) at (0,0) {\frame{\includegraphics[width=\textwidth]{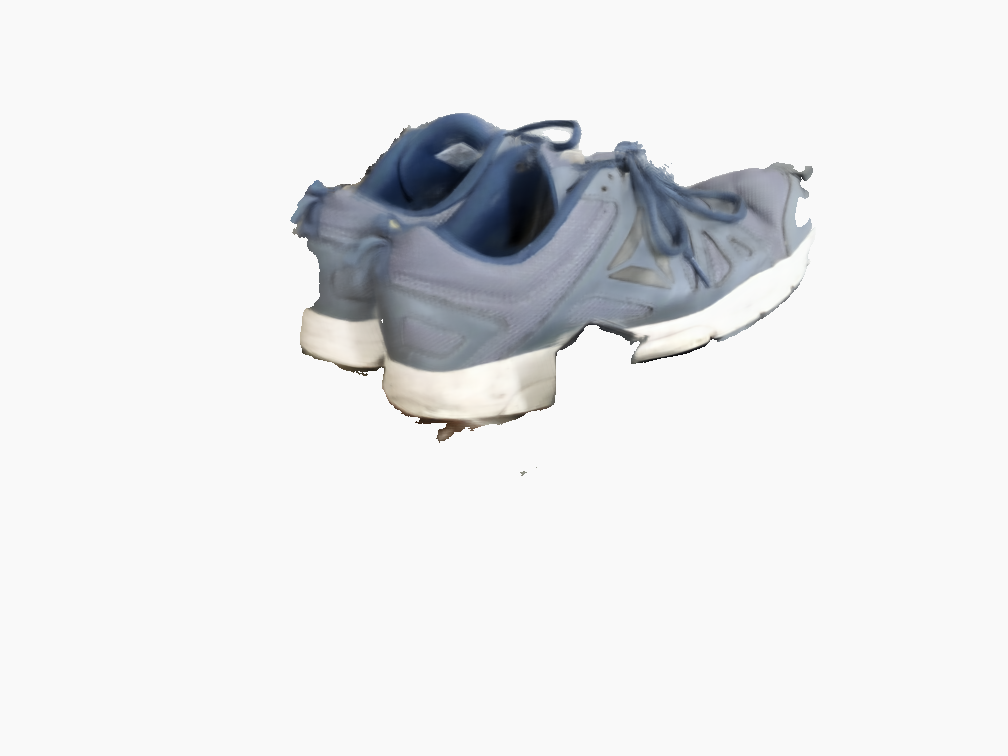}}};
            \begin{scope}[x={(image.south east)},y={(image.north west)}]
                \draw[green,thick] (0.45,0.42) rectangle (0.70, 0.60);
            \end{scope}
        \end{tikzpicture}
    \end{minipage}
    
    \caption{\label{fig:results}{\em Comparison:} Row 1 shows the reference scenes. Row 2 shows the segmentation results of N3F/DFF \cite{N3F, DFF} with the corresponding patch query. Row 3 shows segmentation results of ISRF \cite{isrf} with strokes. Row 4 shows the segmentation results of our GSN method. It is to be noted that the previous methods rely on scene-specific training to enable segmentation. For more details (highlighted boxes), please refer to the Results section in the manuscript.}

\end{figure*}

\section{Method}
\begin{figure*}[!ht]
    \centering
    
    \begin{minipage}{0.33\linewidth}
        \centering
        \frame{\includegraphics[width=\textwidth]{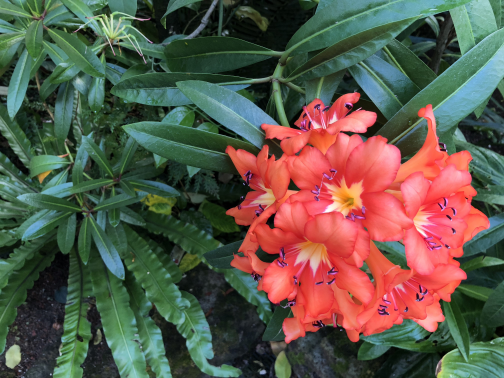}}
    \end{minipage}
    \begin{minipage}{0.33\linewidth}
        \centering
        \frame{\includegraphics[width=\textwidth]{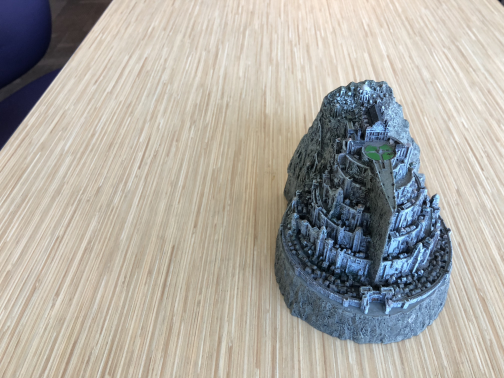}}
    \end{minipage}
    \begin{minipage}{0.33\linewidth}
        \centering
        \frame{\includegraphics[width=\textwidth]{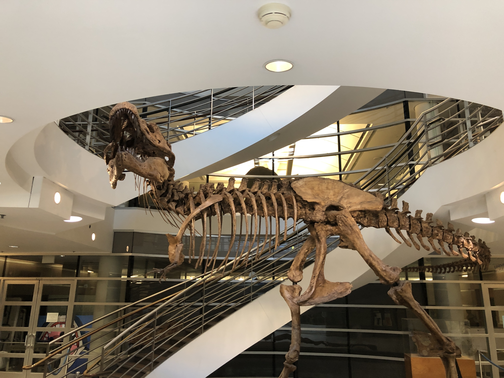}}
    \end{minipage}

    \begin{minipage}{0.33\linewidth}
        \centering
        \frame{\includegraphics[width=\textwidth]{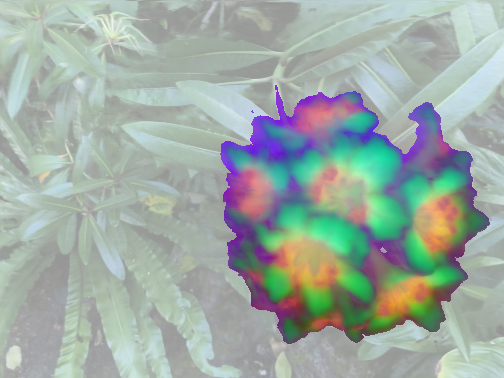}}
    \end{minipage}
    \begin{minipage}{0.33\linewidth}
        \centering
        \frame{\includegraphics[width=\textwidth]{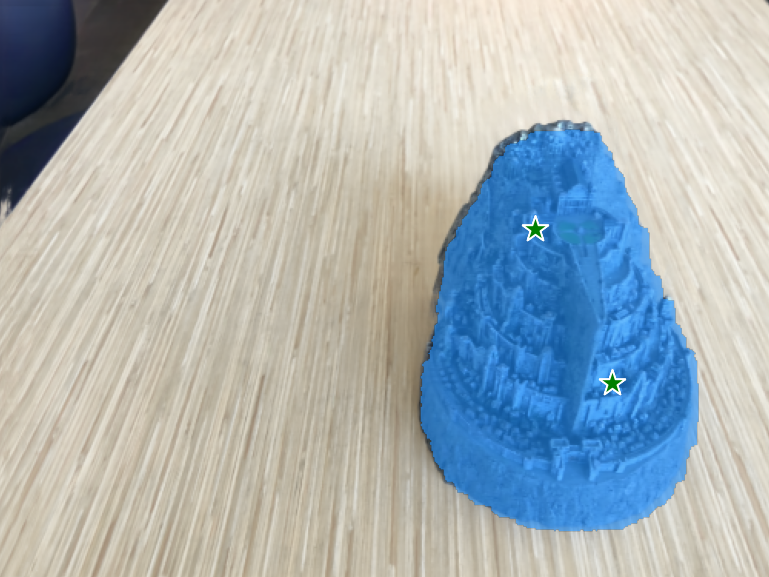}}
    \end{minipage}
    \begin{minipage}{0.33\linewidth}
        \centering
        \frame{\includegraphics[width=\textwidth]{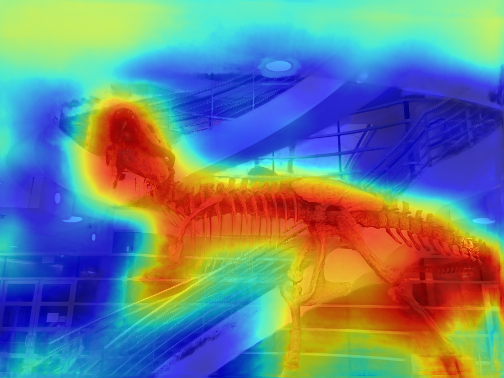}}
    \end{minipage}
    \caption{\label{fig:othersemantics} {\em Other Semantic Fields:} Col. 1 shows the \flower scene with DINOv2 \cite{dinov2} features distilled into it. We show part segmentation of the flowers, i.e., the parts of each flower are coloured the same, depicting the distillation of appropriate features. Col. 2 shows the result on the \fortress scene when SAM \cite{sam} features are distilled into our GSN model, and the SAM decoder is used to segment the image. Col. 3 shows the distillation of CLIP \cite{clip} features into the \trex scene. We use the text-prompt {\em ``a fossil of dinosaur''} to localise the object in the rendered image. The heat map shows how well the pixel corresponds to the text prompt. This figure depicts that our generalised GSN model can incorporate various semantic features.
}
\end{figure*}

{
    \label{sec:method}
    We begin by describing the Generalized NeRF Transformer (GNT) \cite{varma2022attention}. We modify the GNT architecture to aid our semantic feature distillation. We first describe our two-stage training-distillation procedure. Finally, we describe how we perform multi-view segmentation using the distilled features.

    \subsection{Generalised NeRF Transformer}
    {
        \label{sub_sec:method_gnt}
        GNT \cite{varma2022attention} propose to solve the problem of generalization of radiance fields using the following two modules:
        
        \begin{enumerate}
            \item A view transformer aggregates the image features from nearby source views along epipolar lines to predict point-wise features. The role of this module is to interpolate features from nearby views.
            \item A ray transformer accumulates these point-wise features, from the previous module, along a ray. This module acts as a substitute for the volumetric rendering equation. 

        \end{enumerate}
        
        These two modules are stacked one after another sequentially for $b=4$ blocks. The output of the last ray transformer is passed through an MLP (after average pooling), which directly predicts the pixel's colour through which the ray was shot.
        
       The following equations summarise the View-Transformer and Ray-Transformer formulation of GNT:

        \begin{align*}
            \begin{split}
                \mathcal{F}(x, \theta) = V(F_1(\Pi_1(x), \theta), \cdots, F_N(\Pi_N(x), \theta))
                \\
                \mathcal{R}(o, d, \theta) = R(\mathcal{F}({o} + t_1 {d}, {\theta}), \cdots, \mathcal{F}({o} + t_M {d}, {\theta}))
                \\
                {C}({r}) = \operatorname{MLP} \circ \operatorname{Mean} \circ \mathcal{R}(o, d, \theta)
            \end{split}
        \end{align*}
        
        Here, $V(\cdot)$ is the view transformer encoder,  $\Pi_i({x})$ projects position ${x} \in R^3$ onto the $i^{th}$ image plane by applying extrinsic matrix, and ${F}_i({z}, {\theta}) \in {R}^d$ computes the feature vector at position ${z} \in R^2$ and viewing direction $\theta$ via bi-linear interpolation on the feature grids, $\textbf{r} = o + td$ with $t_1, \cdots, t_M$ being uniformly sampled  between near and far plane, $R(\cdot)$ is a standard transformer encoder. $\mathcal{F}$ and $\mathcal{R}$ are the corresponding outputs of the view transformer and ray transformer, respectively, and $C(\cdot)$ is the predicted pixel colour. 
    
    }
    
    \subsection{Modifications to GNT}
    {
        \label{sub_sec:method_modifications}
        
        As explained in the previous section, the Ray-Transformer takes position and view direction as an input parameter. The repeated stacking of the View-Transformer and Ray-Transformer means that the information of view direction is being passed very early to the architecture.
    
        However, as shown by ISRF \cite{isrf}, semantic features are view-agnostic. Looking at an object from a different view direction should not change the semantic features of the object. We remove the view direction as input to the Ray-Transformer to facilitate this. The view-direction input is re-introduced later in the RGB-prediction branch as shown in \cref{fig:modelarch}. Shifting this to a later stage degrades the RGB rendering quality, but very imperceptibly. On the positive side, we obtain a base architecture that generates view-independent feature images, significantly improving segmentation quality. Results related to this decision are shown in the ablation study section.

    }

    \subsection{Feature Distillation}
    {
        \label{sub_sec:method_distillation}
        As shown by \cref{fig:modelarch}, we have a two-stage training mechanism:
        \begin{enumerate}
            \item \textbf{Stage I:} In the first stage, the GNT model with our proposed modification is trained to generate novel views on multiple scenes simultaneously using the RGB Loss $L_{RGB} = \| C_{GT} - C_{Pred}\|^2$.
            \item \textbf{Stage II:} In the second stage, we first obtain pixel-wise features for all the images across all the scenes. We branch off from our direction-independent base model and add another set of view and ray transformers that predict the features. The output features from this student feature head $f_{Pred}$ are compared against the teacher feature images $f_{GT}$ to perform distillation: $L_{Feat.} = \| f_{GT} - f_{Pred} \|^2$.
        \end{enumerate}

    }

    \subsection{Segmentation}
    {
        \label{sub_sec:method_segmentation}
        Similar to N3F, DFF and ISRF, we mainly use DINO \cite{dino} features to perform segmentation. As shown in \cref{fig:othersemantics}, other feature sets may be used.
        
        Given a set of input user strokes, our task is to segment the marked object from all the different views. The preliminary methods like N3F \cite{N3F} propose to use average feature matching to perform segmentation in 3D during the rendering procedure. ISRF \cite{isrf} significantly improves this by first clustering the features marked by the user's stroke and then doing 2D-3D NNFM (Nearest Neighbour Feature Matching).

        We follow a similar strategy as ISRF but in a multi-view setting. The semantic features underlying the user's input stroke are collected and are clustered into $k$ clusters using K-Means clustering. Then, all the pixels in the feature image are classified using the $k$ clusters as to whether they belong to the marked object. This is done using NNFM with all the $k$ clusters. The threshold for the feature distance is identical for all the views and is chosen separately for each scene.

    }
}
\begin{table}[!ht]
\begin{center}
\begin{tabular}{ |c|c|c|c|c| } 
 \hline
 \textbf{Scene} & \textbf{Metric}  & \textbf{N3F} & \textbf{ISRF} & \textbf{Ours} \\ 
 \hline
 \multirow{3}{*}{Chesstable} & Mean IoU $\uparrow$ & 0.344 & \textbf{0.912} & 0.828 \\ \cline{2-5}
 & Accuracy $\uparrow$ & 0.820 & \textbf{0.990} & 0.981\\ \cline{2-5}
 & mAP $\uparrow$ & 0.334 & \textbf{0.916} & 0.833\\ 
 \hline
 
 \multirow{3}{*}{Colorfountain} & Mean IoU $\uparrow$ & 0.871 & \textbf{0.927} & \textbf{0.927} \\ \cline{2-5}
 & Accuracy $\uparrow$ & 0.979 & \textbf{0.989} & \textbf{0.989} \\ \cline{2-5}
 & mAP $\uparrow$ & 0.871 & 0.927 & \textbf{0.93} \\ 
 \hline
 
 \multirow{3}{*}{Stove} & Mean IoU $\uparrow$ & 0.416 & 0.827 & \textbf{0.839} \\ \cline{2-5}
 & Accuracy $\uparrow$ & 0.954 & 0.992 & \textbf{0.993} \\ \cline{2-5}
 & mAP $\uparrow$ & 0.387 & 0.824 & \textbf{0.835} \\ 
 \hline
 
 \multirow{3}{*}{Shoerack} & Mean IoU $\uparrow$ & 0.589 & 0.861 & \textbf{0.911} \\ \cline{2-5}
 & Accuracy $\uparrow$ & 0.913 & 0.980 & \textbf{0.987} \\ \cline{2-5}
 & mAP $\uparrow$ & 0.582 & 0.869 & \textbf{0.912} \\ 
 \hline
\end{tabular}
\end{center}

\caption{\label{tab:llff_metric} This table shows the segmentation metrics against previous works of N3F \cite{N3F}, DFF \cite{DFF}, and ISRF \cite{isrf}. We calculate the mean IoU, accuracy and mean average precision for four scenes from the dataset provided by LLFF \cite{mildenhall2019llff}. Our method performs better than the preliminary works of N3F and DFF while on par with ISRF. The ground truth segmentation masks were hand-annotated for comparison.
}
\vspace{-6mm}
\end{table}



\begin{figure*}[!ht]
    \begin{minipage}{0.32\linewidth}
        \centering
        Original
    \end{minipage}
    \begin{minipage}{0.32\linewidth}
        \centering
        GNT
    \end{minipage}
    \begin{minipage}{0.32\linewidth}
        \centering
        GSN (Ours)
    \end{minipage}
    
    \rotatebox[origin=c]{90}{DINO Features}
    \begin{minipage}{0.32\linewidth}
        \centering
        \begin{tikzpicture}
            \node[anchor=south west,inner sep=0] (image) at (0,0) {\includegraphics[width=\textwidth]{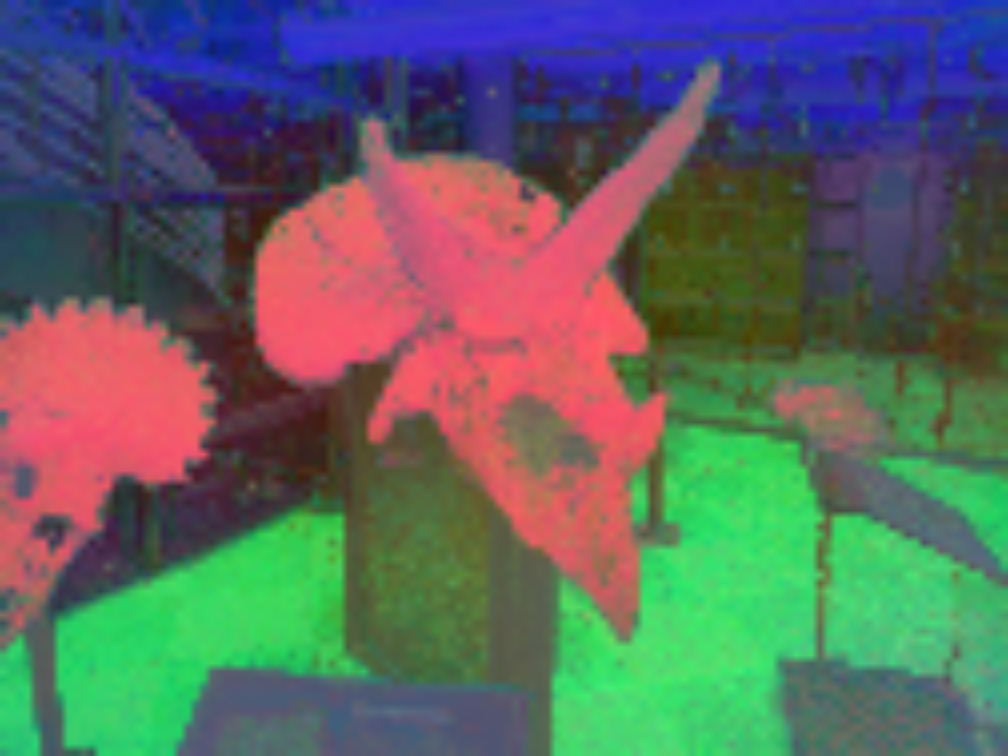}};
            \begin{scope}[x={(image.south east)},y={(image.north west)}]
                \draw[green,thick] (0.01,0.75) rectangle (0.2, 0.9);
                \draw[white,thick] (0.78, 0.1) rectangle (0.85, 0.5);
            \end{scope}
        \end{tikzpicture}
    \end{minipage}
    \begin{minipage}{0.32\linewidth}
        \centering
        \begin{tikzpicture}
            \node[anchor=south west,inner sep=0] (image) at (0,0) {\includegraphics[width=\textwidth]{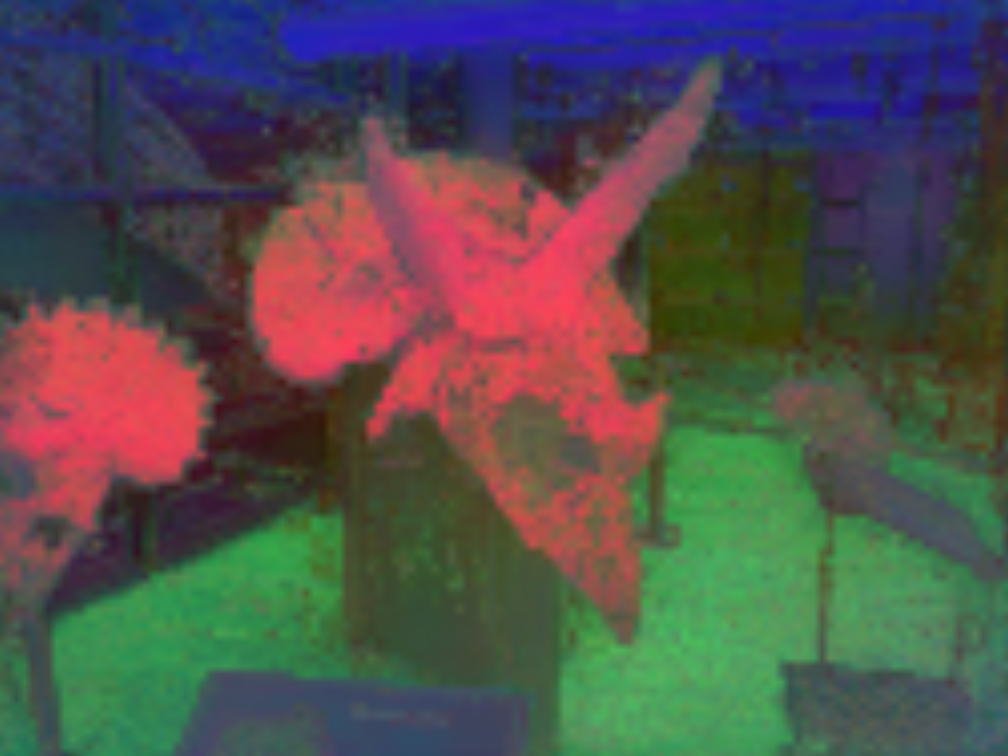}};
            \begin{scope}[x={(image.south east)},y={(image.north west)}]
                \draw[green,thick] (0.01,0.75) rectangle (0.2, 0.9);
                \draw[white,thick] (0.78, 0.1) rectangle (0.85, 0.5);
            \end{scope}
        \end{tikzpicture}
    \end{minipage}
    \begin{minipage}{0.32\linewidth}
        \centering
        \begin{tikzpicture}
            \node[anchor=south west,inner sep=0] (image) at (0,0) {\includegraphics[width=\textwidth]{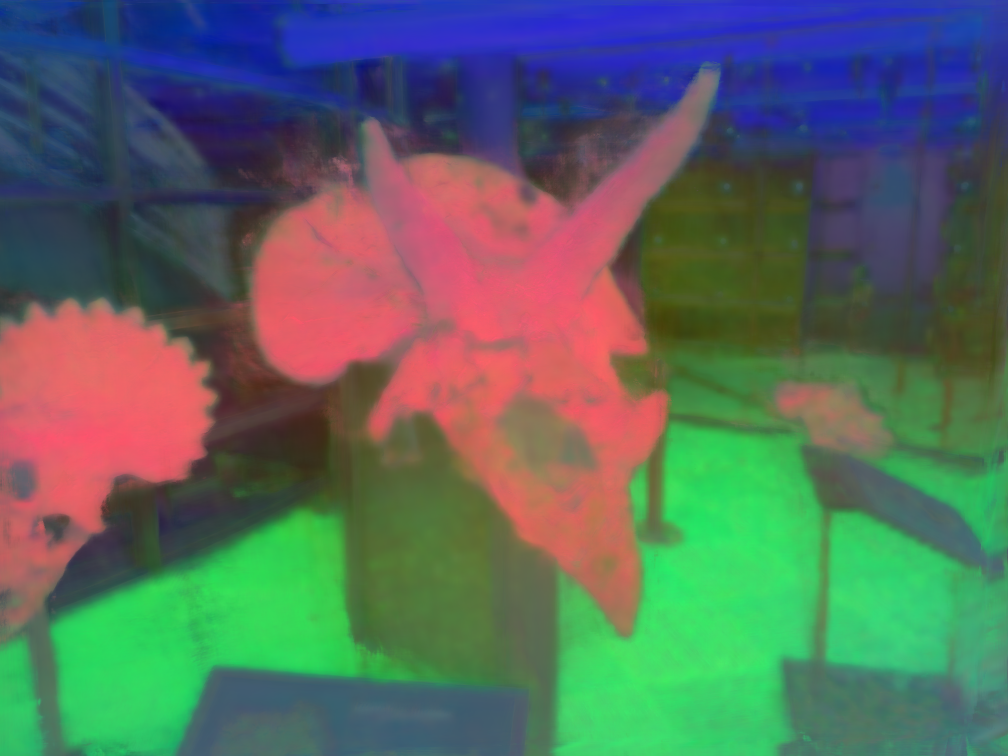}};
            \begin{scope}[x={(image.south east)},y={(image.north west)}]
                \draw[green,thick] (0.01,0.75) rectangle (0.2, 0.9);
                \draw[white,thick] (0.78, 0.1) rectangle (0.85, 0.5);
            \end{scope}
        \end{tikzpicture}
    \end{minipage}

    \rotatebox[origin=c]{90}{Clustered Features}
    \begin{minipage}{0.32\linewidth}
        \centering
        \begin{tikzpicture}
            \node[anchor=south west,inner sep=0] (image) at (0,0) {\includegraphics[width=\textwidth]{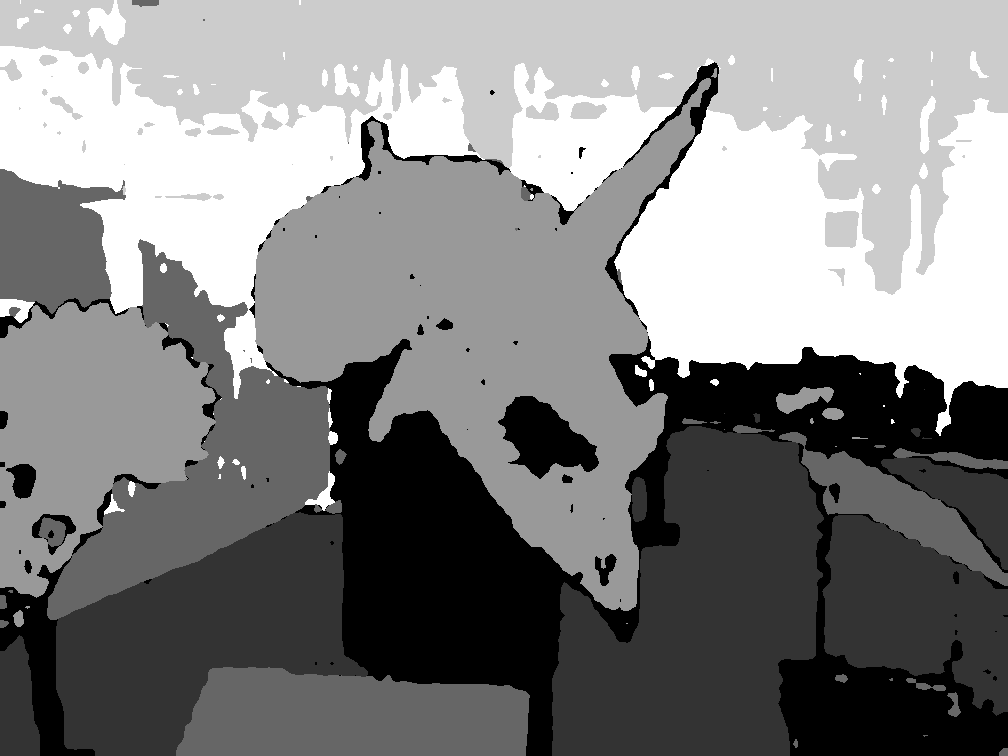}};
            \begin{scope}[x={(image.south east)},y={(image.north west)}]
                \draw[cyan,thick] (0.45,0.3) rectangle (0.65, 0.5);
                \draw[magenta,thick] (0.01, 0.15) rectangle (0.2, 0.6);
            \end{scope}
        \end{tikzpicture}
    \end{minipage}
    \begin{minipage}{0.32\linewidth}
        \centering
        \begin{tikzpicture}
            \node[anchor=south west,inner sep=0] (image) at (0,0) {\includegraphics[width=\textwidth]{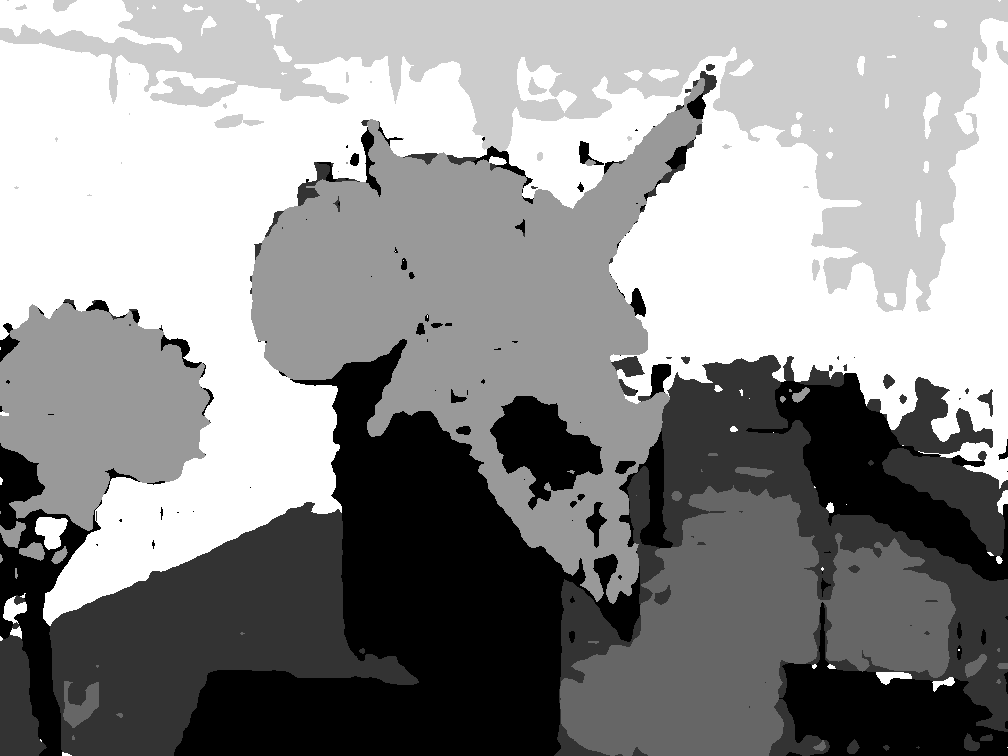}};
            \begin{scope}[x={(image.south east)},y={(image.north west)}]
                \draw[cyan,thick] (0.45,0.3) rectangle (0.65, 0.5);
                \draw[magenta,thick] (0.01, 0.15) rectangle (0.2, 0.6);
            \end{scope}
        \end{tikzpicture}
    \end{minipage}
    \begin{minipage}{0.32\linewidth}
        \centering
        \begin{tikzpicture}
            \node[anchor=south west,inner sep=0] (image) at (0,0) {\includegraphics[width=\textwidth]{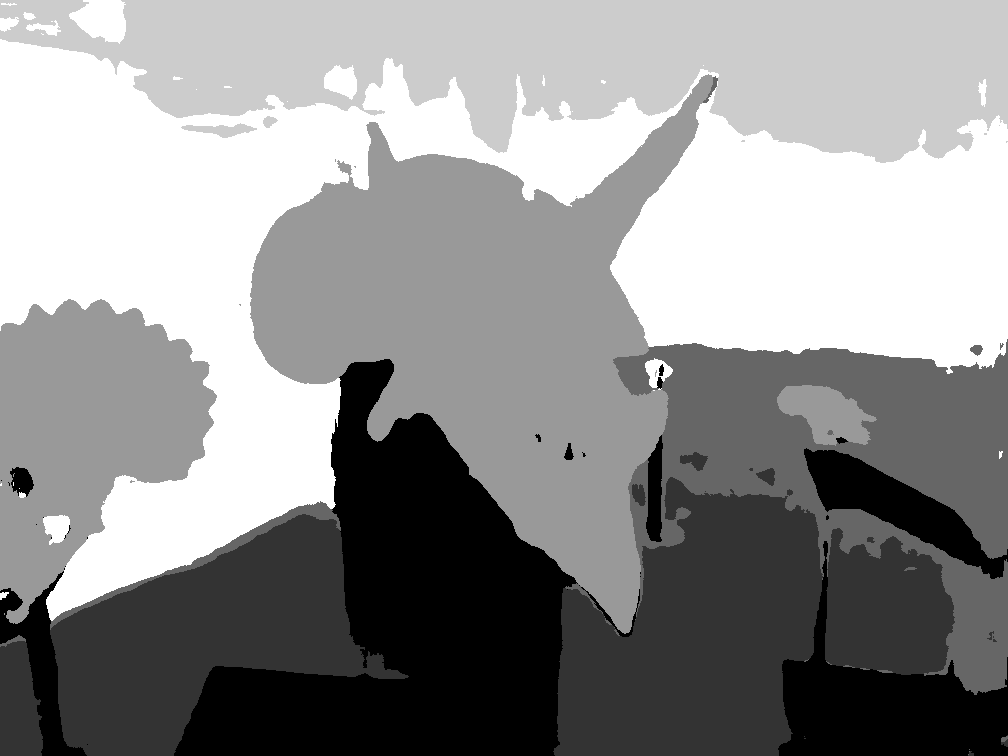}};
            \begin{scope}[x={(image.south east)},y={(image.north west)}]
                \draw[cyan,thick] (0.45,0.3) rectangle (0.65, 0.5);
                \draw[magenta,thick] (0.01, 0.15) rectangle (0.2, 0.6);
            \end{scope}
        \end{tikzpicture}
    \end{minipage}
    
    \caption{\label{fig:better_than_dino} Row 1 shows DINO \cite{dino} features computed in various settings on \horns from LLFF \cite{mildenhall2019llff}. For visual simplification, a 3-dimensional PCA has been done on the features. Col. 1 and Col. 2 show DINO features computed on the original image and on GNT's \cite{varma2022attention} output respectively. Col. 3 shows features predicted by our GSN method. The boxes highlight clear feature differences. We demonstrate better feature quality by doing K-Means Clustering on the feature images as shown in Row 2. Our method gives clear, noise-free clusters as shown by the boxes.}

\end{figure*}

\section{Implementation Details}
{
    \label{sec:implementation_details}

    We use the code provided by \cite{varma2022attention} and implement our methods on top of it. We use 2 RTX 3090 GPUs for distributed training and conducting our experiments. For extraction of features, we run PCA on every image on the scene to reduce the feature dimension to 64 and normalise these features. For Stage I training, we use four blocks of view-ray transformers and one block of view-ray transformer in Stage II training. We used 512 rays, 192 points and 200,000 iterations for Stage-I and trained it for two days. Stage II is trained only for 5,000 iterations with 512 rays and 192 points and trained only for 4 hours. We use a learning rate of 1e-3 for the Stage II training and decrease the learning rate of Stage-I by a factor of 10 during the Stage II training. The weight factor of the RGB loss $L_{RGB}$ is set to 0.1, and feature loss $L_{Feat}$ is set to 1 during Stage II training. We select 10 source views for every target view, and the training is done at an image resolution of 756x1008. It takes around 60 seconds to render an image. The dataset provided by IBRNet \cite{ibrnet} and some scenes from the LLFF Dataset \cite{mildenhall2019llff} are used for training our generalised model. For testing, we use the LLFF \cite{mildenhall2019llff} dataset, which was unseen during the training phase. These images are passed through DINO ViT-b8 to obtain feature images. The number of clusters for segmentation is set to $11$, although a slight variation does not make a difference.
    
}

\vspace{-2mm}
\section{Results}
{
    \label{sec:results}
    
    This section shows qualitative and quantitative results against previous works that perform multi-view segmentation using radiance fields. We demonstrate that our work improves the feature prediction while at the same time generalising by interpolating the features across multiple views. Our method can generalise to any set of features, and we show this by incorporating DINO \cite{dino}, CLIP \cite{clip}, SAM \cite{sam} and DINOv2 \cite{dinov2} features.  
    
    \textbf{Note:} Every result (for our method) shown in this paper/supplementary is a scene not present in the training set, i.e. a previously unseen scene.

    \subsection{Comparison}
    {
        \label{sub_sec:results_comparison}
        
        We compare our results against the previous works of N3F \cite{N3F}, DFF \cite{DFF}, and ISRF \cite{isrf}. N3F and DFF use similar methods for their segmentation, so we treat them as one method as done by previous works \cite{isrf}. These works utilise feature distillation from 2D feature images into a 3D radiance field and use these distilled features to perform semantic segmentation. All these works require scene-specific training before segmentation can be achieved, while we propose a generalised framework. Despite this disadvantage, we perform on par with ISRF and better than N3F/DFF.

        \cref{fig:results} shows our segmentation results against N3F/DFF and ISRF. Due to an average feature matching method, the results of N3F/DFF are filled with noise alongside occasional bleeding-in of unmarked objects (\chesstable). ISRF uses K-Means Clustering followed by NNFM to improve the results significantly. K-Means Clustering helps in reducing the noise, and NNFM helps in the selection of the correct object. As explained before, we follow a similar methodology in 2D to achieve segmentation.

        \cref{tab:llff_metric} shows quantitative results on four scenes taken from the LLFF \cite{mildenhall2019llff} dataset. We calculate {\em meanIoU}, {\em accuracy} and {\em mean average precision score} on four scenes from the LLFF \cite{mildenhall2019llff} dataset. These metrics show that our method is better than N3F/DFF and comes close to the state-of-the-art method ISRF and, in some cases, even performs better than them.

        Please refer to the highlighted boxes in \cref{fig:results} for the following discussion. In \chesstable scene, ISRF performs better than us. The area below the table is not segmented well because of fewer views of that area. Our method heavily relies on appropriate epipolar geometry prediction, and fewer views lead to bad epipolar geometry. Similarly, in \colorfountain scene, our segmentation suffers under the basin due to fewer views. ISRF cleverly counters this using their 3D region growing. In the \stove scene, we are able to recover more content than ISRF. In the \shoerack scene, the white sole is recovered more in our case than compared to ISRF. An important observation needs to be noted. It can be seen that the N3F/DFF and ISRF show the part of the shoe that is occluded by the shoerack because they are doing 3D segmentation. Since we are doing multi-view segmentation instead of 3D segmentation, the occluder is negated in our result.

    }

    \subsection{Student Surpasses Teacher}
    {
        \label{sub_sec:results_surpass}
        There are three methods to obtain a good semantic feature representation for a given set of multi-view images. The first method is to directly pass the image through a feature extractor and get the features. If one wants features from an unseen view, one can use novel view synthesis methods on a scene \cite{varma2022attention,nerf} and then pass the novel view through a feature extractor. We show results on these methods in Fig. \ref{fig:better_than_dino} and compare them against our approach, which also produces features on novel views and that too in a generalised fashion. The figure shows the fine-grained features that we produce. Essentially, our method can interpolate the coarse features from different views and combine them with the learnt geometry to produce fine-grained features. Our student GSN method can surpass the teacher feature extractor methods to produce better features in a multi-view setting. The feature images shown in the figure are obtained by doing a PCA of the features to 3 dimensions which correspond to RGB channels of the image. For results on more scenes, please refer to the supplementary document.

    }

    \subsection{Other Semantic Fields}
    {
        \label{sub_sec:results_other_fields}
        The generalised feature distillation we propose can be applied to any set of features. We demonstrate this by distilling three different features using our method. We have already shown results on DINO\cite{dino} in \cref{fig:results}. In \cref{fig:othersemantics}, we show the distillation of DINOv2 \cite{dinov2} features, CLIP \cite{clip} features, and SAM \cite{sam} features.
        We show part-segmentation using DINOv2, segmentation using CLIP space using a text prompt and segmentation using SAM features space by passing them through the SAM decoder. These results indicate that our generalisation approach can be adapted to any feature space by fine-tuning Stage II. For more details on the exact method, please refer to supplementary.

    }
}
\begin{figure}[bt]
    \centering
    \begin{minipage}{0.49\linewidth}
    \centering
    \frame{\includegraphics[width=\textwidth]{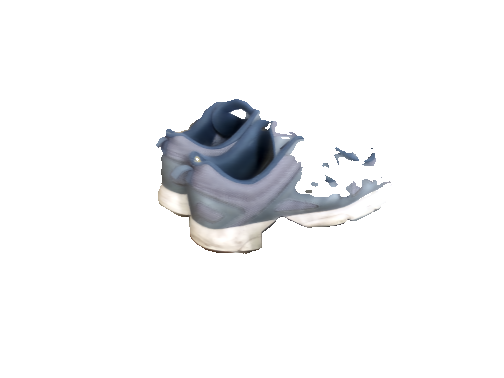}}
    \end{minipage}
    \begin{minipage}{0.49\linewidth}
    \centering
    \frame{\includegraphics[width=\textwidth]{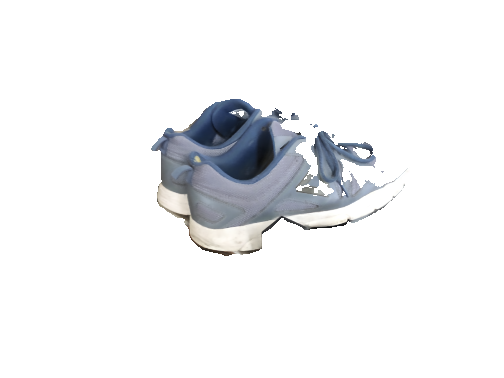}}
    \end{minipage}
    
    \caption{\label{fig:ablation_view_direction} Left and right images show segmentation results with original GNT and our GSN architectures respectively. We input the same single stroke and threshold for segmentation. In our case, the features are more coherent, leading to more accurate segmentation.
}
 
\end{figure}
\begin{figure}[!ht]
    \centering
    
    \begin{minipage}{0.49\linewidth}
    \centering
    \frame{\includegraphics[width=\textwidth]{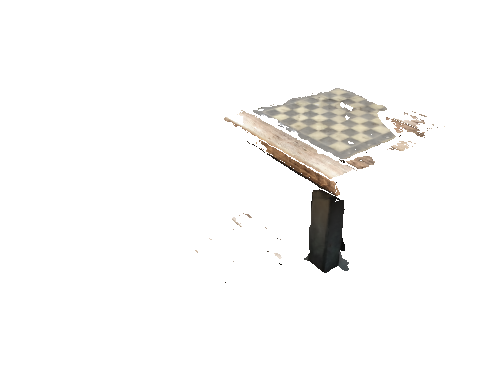}}
    \end{minipage}
    \begin{minipage}{0.49\linewidth}
    \centering
    \frame{\includegraphics[width=\textwidth]{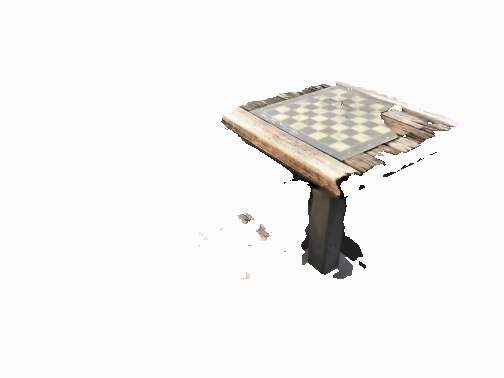}}
    \end{minipage}
    
    \caption{\label{fig:ablation_vol_render} The left and right images show the segmentation result when we using volumetric rendering and ray transformer (our method) respectively. In both cases, we tweak the threshold of segmentation to the point such that the chair (left of table) barely starts to bleed in. In case of ray transformer, results are improved indicating that it is a better feature aggregator.
}
 
\end{figure}
\section{Ablation Study}
{
    \label{sec:ablation}
    \subsection{Architecture Modification}
    {
        \label{sub_sec:ablation_architecture}
        An essential modification to the original GNT architecture is displacing the view directions as input. We train the original GNT architecture and attempt to do segmentation with it using one user stroke. \cref{fig:ablation_view_direction} shows the segmentation results obtained on the original GNT model and with our modification when the same user stroke is provided. In our case, the segmented object covers a more significant part of the underlying object. In the case of the original GNT architecture, the semantic features are conditioned upon the view directions, which is inherently wrong since semantic features are view-agnostic. In common words, they are diffuse and not specular. This leads to incorrect feature distillation and poorer segmentation results in GNT architecture.

    }

    \subsection{Volumteric Rendering}
    {
        \label{sub_sec:ablation_volumetric}


        One may suspect that the volumetric-rendering variant mentioned in GNT \cite{varma2022attention} should give better results since it performs segmentation in 3D. We test this with our GSN method and observe that the results are subpar to our method, as shown in \cref{fig:ablation_vol_render}. This indicates that the ray-transformer blocks are better aggregators for image-based rendering methods than volumetric rendering equation for density, colour and feature values.

    }
}

\section{Conclusion}
{
    \label{sec:conclusion}
    
    We present a novel method for multi-view segmentation. The essential advantage of our approach is its generalisability, i.e. it can perform segmentation on arbitrarily new scenes without any training. This differentiates it from the previous methods. We compare our results against the earlier methods and show that we perform at par with them while being generalisable to unseen scenes. This is a big step in bringing the applications of generalisable neural radiance fields closer to scene-specific radiance fields. The features predicted by our method can be used for several downstream tasks.
    
    \paragraph{Limitations \& Future Work. } Since we rely on a transformer-based architecture, the rendering process is inherently slow compared to the several scene-specific radiance field methods. Improving the rendering speed can significantly improve the human interaction experience required for our stroke-based segmentation method. We leave the rendering speed improvement of generalisable radiance fields as future work. Currently, our method performs multi-view segmentation since it uses image-based rendering. 
    Some applications require 3D segmentation instead of multi-view segmentation. 
    Thus, a generalisable 3D segmentation framework is promising for future work.

}
\section*{Acknowledgements}
{
    We thank Prof Kaushik Mitra of IIT Madras, for advice and computational support. We also thank Mukund Varma of UCSD for discussions and insights on GNT.
}

\bibliography{main}

\begin{thebibliography}{42}
\providecommand{\natexlab}[1]{#1}

\bibitem[{Agarwal et~al.(2009)Agarwal, Snavely, Simon, Seitz, and
  Szeliski}]{romein1day}
Agarwal, S.; Snavely, N.; Simon, I.; Seitz, S.~M.; and Szeliski, R. 2009.
\newblock Building Rome in a day.
\newblock In \emph{Proceedings of the IEEE/CVF Conference on Computer Vision
  and Pattern Recognition (CVPR)}.

\bibitem[{Caron et~al.(2021)Caron, Touvron, Misra, J\'egou, Mairal, Bojanowski,
  and Joulin}]{dino}
Caron, M.; Touvron, H.; Misra, I.; J\'egou, H.; Mairal, J.; Bojanowski, P.; and
  Joulin, A. 2021.
\newblock {Emerging Properties in Self-Supervised Vision Transformers}.
\newblock In \emph{Proceedings of the IEEE/CVF International Conference on
  Computer Vision (ICCV)}.

\bibitem[{Chen et~al.(2022{\natexlab{a}})Chen, Xu, Geiger, Yu, and
  Su}]{tensorf}
Chen, A.; Xu, Z.; Geiger, A.; Yu, J.; and Su, H. 2022{\natexlab{a}}.
\newblock {TensoRF: Tensorial Radiance Fields}.
\newblock In \emph{Proceedings of the European Conference on Computer Vision
  (ECCV)}.

\bibitem[{Chen et~al.(2021)Chen, Xu, Zhao, Zhang, Xiang, Yu, and Su}]{MVSNeRF}
Chen, A.; Xu, Z.; Zhao, F.; Zhang, X.; Xiang, F.; Yu, J.; and Su, H. 2021.
\newblock Mvsnerf: Fast generalizable radiance field reconstruction from
  multi-view stereo.
\newblock In \emph{Proceedings of the IEEE/CVF International Conference on
  Computer Vision (ICCV)}.

\bibitem[{Chen et~al.(2022{\natexlab{b}})Chen, Zhang, Li, Chen, Feng, Wang, and
  Wang}]{chen2022hallucinated}
Chen, X.; Zhang, Q.; Li, X.; Chen, Y.; Feng, Y.; Wang, X.; and Wang, J.
  2022{\natexlab{b}}.
\newblock Hallucinated neural radiance fields in the wild.
\newblock In \emph{Proceedings of the IEEE/CVF Conference on Computer Vision
  and Pattern Recognition}.

\bibitem[{Cherti et~al.(2023)Cherti, Beaumont, Wightman, Wortsman, Ilharco,
  Gordon, Schuhmann, Schmidt, and Jitsev}]{cherti2023reproducible}
Cherti, M.; Beaumont, R.; Wightman, R.; Wortsman, M.; Ilharco, G.; Gordon, C.;
  Schuhmann, C.; Schmidt, L.; and Jitsev, J. 2023.
\newblock Reproducible scaling laws for contrastive language-image learning.
\newblock In \emph{Proceedings of the IEEE/CVF Conference on Computer Vision
  and Pattern Recognition}.

\bibitem[{Fridovich-Keil et~al.(2023)Fridovich-Keil, Meanti, Warburg, Recht,
  and Kanazawa}]{kplanes}
Fridovich-Keil, S.; Meanti, G.; Warburg, F.~R.; Recht, B.; and Kanazawa, A.
  2023.
\newblock K-Planes: Explicit Radiance Fields in Space, Time, and Appearance.
\newblock In \emph{Proceedings of the IEEE/CVF Conference on Computer Vision
  and Pattern Recognition (CVPR)}.

\bibitem[{{Fridovich-Keil and Yu} et~al.(2022){Fridovich-Keil and Yu}, Tancik,
  Chen, Recht, and Kanazawa}]{plenoxels}
{Fridovich-Keil and Yu}; Tancik, M.; Chen, Q.; Recht, B.; and Kanazawa, A.
  2022.
\newblock {Plenoxels: Radiance Fields without Neural Networks}.
\newblock In \emph{Proceedings of the IEEE/CVF Conference on Computer Vision
  and Pattern Recognition (CVPR)}.

\bibitem[{Goel et~al.(2023)Goel, Sirikonda, Saini, and Narayanan}]{isrf}
Goel, R.; Sirikonda, D.; Saini, S.; and Narayanan, P. 2023.
\newblock {Interactive Segmentation of Radiance Fields}.
\newblock In \emph{{Proceedings of the IEEE/CVF Conference on Computer Vision
  and Pattern Recognition (CVPR)}}.

\bibitem[{Huang et~al.(2022)Huang, He, Yuan, Lai, and Gao}]{stylizednerf_cvpr}
Huang, Y.-H.; He, Y.; Yuan, Y.-J.; Lai, Y.-K.; and Gao, L. 2022.
\newblock {StylizedNeRF: Consistent 3D Scene Stylization as Stylized NeRF via
  2D-3D Mutual Learning}.
\newblock In \emph{Proceedings of the IEEE/CVF Conference on Computer Vision
  and Pattern Recognition (CVPR)}.

\bibitem[{Kerbl et~al.(2023)Kerbl, Kopanas, Leimk{\"u}hler, and
  Drettakis}]{gsplat}
Kerbl, B.; Kopanas, G.; Leimk{\"u}hler, T.; and Drettakis, G. 2023.
\newblock 3D Gaussian Splatting for Real-Time Radiance Field Rendering.
\newblock \emph{ACM Transactions on Graphics}.

\bibitem[{Kerr et~al.(2023)Kerr, Kim, Goldberg, Kanazawa, and Tancik}]{lerf}
Kerr, J.; Kim, C.~M.; Goldberg, K.; Kanazawa, A.; and Tancik, M. 2023.
\newblock LERF: Language Embedded Radiance Fields.
\newblock In \emph{International Conference on Computer Vision (ICCV)}.

\bibitem[{Kirillov et~al.(2023)Kirillov, Mintun, Ravi, Mao, Rolland, Gustafson,
  Xiao, Whitehead, Berg, Lo, Doll{\'a}r, and Girshick}]{sam}
Kirillov, A.; Mintun, E.; Ravi, N.; Mao, H.; Rolland, C.; Gustafson, L.; Xiao,
  T.; Whitehead, S.; Berg, A.~C.; Lo, W.-Y.; Doll{\'a}r, P.; and Girshick, R.
  2023.
\newblock Segment Anything.
\newblock \emph{arXiv:2304.02643}.

\bibitem[{Kobayashi, Matsumoto, and Sitzmann(2022)}]{DFF}
Kobayashi, S.; Matsumoto, E.; and Sitzmann, V. 2022.
\newblock {Decomposing NeRF for Editing via Feature Field Distillation}.
\newblock In \emph{Adv. Neural Inform. Process. Syst.}

\bibitem[{Li et~al.(2022)Li, Weinberger, Belongie, Koltun, and Ranftl}]{lseg}
Li, B.; Weinberger, K.~Q.; Belongie, S.; Koltun, V.; and Ranftl, R. 2022.
\newblock {Language-driven Semantic Segmentation}.
\newblock In \emph{Int. Conf. Learn. Represent.}

\bibitem[{Martin-Brualla et~al.(2021)Martin-Brualla, Radwan, Sajjadi, Barron,
  Dosovitskiy, and Duckworth}]{martinbrualla2020nerfw}
Martin-Brualla, R.; Radwan, N.; Sajjadi, M. S.~M.; Barron, J.~T.; Dosovitskiy,
  A.; and Duckworth, D. 2021.
\newblock {NeRF in the Wild: Neural Radiance Fields for Unconstrained Photo
  Collections}.
\newblock In \emph{CVPR}.

\bibitem[{Mildenhall et~al.(2019)Mildenhall, Srinivasan, Ortiz-Cayon,
  Kalantari, Ramamoorthi, Ng, and Kar}]{mildenhall2019llff}
Mildenhall, B.; Srinivasan, P.~P.; Ortiz-Cayon, R.; Kalantari, N.~K.;
  Ramamoorthi, R.; Ng, R.; and Kar, A. 2019.
\newblock {Local Light Field Fusion: Practical View Synthesis with Prescriptive
  Sampling Guidelines}.
\newblock \emph{ACM Trans. Graph.}

\bibitem[{Mildenhall et~al.(2020)Mildenhall, Srinivasan, Tancik, Barron,
  Ramamoorthi, and Ng}]{nerf}
Mildenhall, B.; Srinivasan, P.~P.; Tancik, M.; Barron, J.~T.; Ramamoorthi, R.;
  and Ng, R. 2020.
\newblock {NeRF: Representing Scenes as Neural Radiance Fields for View
  Synthesis}.
\newblock In \emph{Proceedings of the European Conference on Computer Vision
  (ECCV)}.

\bibitem[{Mirzaei et~al.(2023)Mirzaei, Aumentado-Armstrong, Derpanis, Kelly,
  Brubaker, Gilitschenski, and Levinshtein}]{spinnerf}
Mirzaei, A.; Aumentado-Armstrong, T.; Derpanis, K.~G.; Kelly, J.; Brubaker,
  M.~A.; Gilitschenski, I.; and Levinshtein, A. 2023.
\newblock {SPIn-NeRF}: Multiview Segmentation and Perceptual Inpainting with
  Neural Radiance Fields.
\newblock In \emph{CVPR}.

\bibitem[{M\"uller et~al.(2022)M\"uller, Evans, Schied, and
  Keller}]{instantngp}
M\"uller, T.; Evans, A.; Schied, C.; and Keller, A. 2022.
\newblock {Instant Neural Graphics Primitives with a Multiresolution Hash
  Encoding}.
\newblock \emph{ACM Trans. Graph.}

\bibitem[{Narayanan, Rander, and Kanade(1998)}]{PJN:ICCV98}
Narayanan, P.~J.; Rander, P.~W.; and Kanade, T. 1998.
\newblock {Constructing Virtual Worlds Using Dense Stereo}.
\newblock In \emph{Proceedings of the IEEE/CVF International Conference on
  Computer Vision (ICCV)}.

\bibitem[{Niemeyer et~al.(2022)Niemeyer, Barron, Mildenhall, Sajjadi, Geiger,
  and Radwan}]{Niemeyer2021Regnerf}
Niemeyer, M.; Barron, J.~T.; Mildenhall, B.; Sajjadi, M. S.~M.; Geiger, A.; and
  Radwan, N. 2022.
\newblock RegNeRF: Regularizing Neural Radiance Fields for View Synthesis from
  Sparse Inputs.
\newblock In \emph{Proc. IEEE Conf. on Computer Vision and Pattern Recognition
  (CVPR)}.

\bibitem[{Oquab et~al.(2023)Oquab, Darcet, Moutakanni, Vo, Szafraniec,
  Khalidov, Fernandez, Haziza, Massa, El-Nouby, Howes, Huang, Xu, Sharma, Li,
  Galuba, Rabbat, Assran, Ballas, Synnaeve, Misra, Jegou, Mairal, Labatut,
  Joulin, and Bojanowski}]{dinov2}
Oquab, M.; Darcet, T.; Moutakanni, T.; Vo, H.~V.; Szafraniec, M.; Khalidov, V.;
  Fernandez, P.; Haziza, D.; Massa, F.; El-Nouby, A.; Howes, R.; Huang, P.-Y.;
  Xu, H.; Sharma, V.; Li, S.-W.; Galuba, W.; Rabbat, M.; Assran, M.; Ballas,
  N.; Synnaeve, G.; Misra, I.; Jegou, H.; Mairal, J.; Labatut, P.; Joulin, A.;
  and Bojanowski, P. 2023.
\newblock DINOv2: Learning Robust Visual Features without Supervision.
\newblock \emph{arXiv:2304.07193}.

\bibitem[{Park et~al.(2021)Park, Sinha, Hedman, Barron, Bouaziz, Goldman,
  Martin-Brualla, and Seitz}]{park2021hypernerf}
Park, K.; Sinha, U.; Hedman, P.; Barron, J.~T.; Bouaziz, S.; Goldman, D.~B.;
  Martin-Brualla, R.; and Seitz, S.~M. 2021.
\newblock HyperNeRF: A Higher-Dimensional Representation for Topologically
  Varying Neural Radiance Fields.
\newblock \emph{ACM Trans. Graph.}

\bibitem[{Radford et~al.(2021)Radford, Kim, Hallacy, Ramesh, Goh, Agarwal,
  Sastry, Askell, Mishkin, Clark, Krueger, and Sutskever}]{clip}
Radford, A.; Kim, J.~W.; Hallacy, C.; Ramesh, A.; Goh, G.; Agarwal, S.; Sastry,
  G.; Askell, A.; Mishkin, P.; Clark, J.; Krueger, G.; and Sutskever, I. 2021.
\newblock Learning Transferable Visual Models From Natural Language
  Supervision.
\newblock In \emph{International Conference on Machine Learning}.

\bibitem[{Ren et~al.(2022)Ren, Agarwala, Russell, Schwing, and Wang}]{nvos}
Ren, Z.; Agarwala, A.; Russell, B.; Schwing, A.~G.; and Wang, O. 2022.
\newblock {Neural Volumetric Object Selection}.
\newblock In \emph{Proceedings of the IEEE/CVF Conference on Computer Vision
  and Pattern Recognition (CVPR)}.

\bibitem[{Snavely, Seitz, and Szeliski(2006)}]{phototourism}
Snavely, N.; Seitz, S.~M.; and Szeliski, R. 2006.
\newblock Photo Tourism: Exploring Photo Collections in 3D.
\newblock \emph{ACM Trans. Graph.}

\bibitem[{Sun, Sun, and Chen(2022)}]{dvgo}
Sun, C.; Sun, M.; and Chen, H. 2022.
\newblock {Direct Voxel Grid Optimization: Super-fast Convergence for Radiance
  Fields Reconstruction}.
\newblock In \emph{Proceedings of the IEEE/CVF Conference on Computer Vision
  and Pattern Recognition (CVPR)}.

\bibitem[{Tancik et~al.(2022)Tancik, Casser, Yan, Pradhan, Mildenhall,
  Srinivasan, Barron, and Kretzschmar}]{tancik2022block}
Tancik, M.; Casser, V.; Yan, X.; Pradhan, S.; Mildenhall, B.; Srinivasan,
  P.~P.; Barron, J.~T.; and Kretzschmar, H. 2022.
\newblock Block-nerf: Scalable large scene neural view synthesis.
\newblock In \emph{Proceedings of the IEEE/CVF Conference on Computer Vision
  and Pattern Recognition}.

\bibitem[{Tewari et~al.(2022)Tewari, Thies, Mildenhall, Srinivasan, Tretschk,
  Wang, Lassner, Sitzmann, Martin-Brualla, Lombardi, Simon, Theobalt, Nießner,
  Barron, Wetzstein, Zollhöfer, and Golyanik}]{nerf_survey}
Tewari, A.; Thies, J.; Mildenhall, B.; Srinivasan, P.; Tretschk, E.; Wang, Y.;
  Lassner, C.; Sitzmann, V.; Martin-Brualla, R.; Lombardi, S.; Simon, T.;
  Theobalt, C.; Nießner, M.; Barron, J.~T.; Wetzstein, G.; Zollhöfer, M.; and
  Golyanik, V. 2022.
\newblock {Advances in Neural Rendering}.
\newblock \emph{Comput. Graph. Forum}.

\bibitem[{Tschernezki et~al.(2022)Tschernezki, Laina, Larlus, and
  Vedaldi}]{N3F}
Tschernezki, V.; Laina, I.; Larlus, D.; and Vedaldi, A. 2022.
\newblock {Neural Feature Fusion Fields}: {3D} Distillation of Self-Supervised
  {2D} Image Representations.
\newblock In \emph{International Conference on 3D Vision (3DV)}.

\bibitem[{Turki, Ramanan, and Satyanarayanan(2022)}]{Turki_2022_CVPR}
Turki, H.; Ramanan, D.; and Satyanarayanan, M. 2022.
\newblock Mega-NERF: Scalable Construction of Large-Scale NeRFs for Virtual
  Fly-Throughs.
\newblock In \emph{Proceedings of the IEEE/CVF Conference on Computer Vision
  and Pattern Recognition (CVPR)}.

\bibitem[{Varma et~al.(2023)Varma, Wang, Chen, Chen, Venugopalan, and
  Wang}]{varma2022attention}
Varma, M.; Wang, P.; Chen, X.; Chen, T.; Venugopalan, S.; and Wang, Z. 2023.
\newblock Is Attention All That NeRF Needs?
\newblock In \emph{The Eleventh International Conference on Learning
  Representations}.

\bibitem[{Verbin et~al.(2022)Verbin, Hedman, Mildenhall, Zickler, Barron, and
  Srinivasan}]{verbin2022refnerf}
Verbin, D.; Hedman, P.; Mildenhall, B.; Zickler, T.; Barron, J.~T.; and
  Srinivasan, P.~P. 2022.
\newblock {Ref-NeRF}: Structured View-Dependent Appearance for Neural Radiance
  Fields.
\newblock \emph{CVPR}.

\bibitem[{Wang et~al.(2022)Wang, Chai, He, Chen, and Liao}]{wang2022clip}
Wang, C.; Chai, M.; He, M.; Chen, D.; and Liao, J. 2022.
\newblock Clip-nerf: Text-and-image driven manipulation of neural radiance
  fields.
\newblock In \emph{Proceedings of the IEEE/CVF Conference on Computer Vision
  and Pattern Recognition}.

\bibitem[{Wang et~al.(2021{\natexlab{a}})Wang, Liu, Liu, Theobalt, Komura, and
  Wang}]{neus}
Wang, P.; Liu, L.; Liu, Y.; Theobalt, C.; Komura, T.; and Wang, W.
  2021{\natexlab{a}}.
\newblock NeuS: Learning Neural Implicit Surfaces by Volume Rendering for
  Multi-view Reconstruction.
\newblock \emph{NeurIPS}.

\bibitem[{Wang et~al.(2021{\natexlab{b}})Wang, Wang, Genova, Srinivasan, Zhou,
  Barron, Martin-Brualla, Snavely, and Funkhouser}]{ibrnet}
Wang, Q.; Wang, Z.; Genova, K.; Srinivasan, P.~P.; Zhou, H.; Barron, J.~T.;
  Martin-Brualla, R.; Snavely, N.; and Funkhouser, T. 2021{\natexlab{b}}.
\newblock IBRNet: Learning Multi-View Image-Based Rendering.
\newblock In \emph{Proceedings of the IEEE/CVF Conference on Computer Vision
  and Pattern Recognition (CVPR)}.

\bibitem[{Xie et~al.(2022)Xie, Takikawa, Saito, Litany, Yan, Khan, Tombari,
  Tompkin, Sitzmann, and Sridhar}]{nerf_survey2}
Xie, Y.; Takikawa, T.; Saito, S.; Litany, O.; Yan, S.; Khan, N.; Tombari, F.;
  Tompkin, J.; Sitzmann, V.; and Sridhar, S. 2022.
\newblock {Neural Fields in Visual Computing and Beyond}.
\newblock \emph{Comput. Graph. Forum}.

\bibitem[{Xu et~al.(2022)Xu, Jiang, Wang, Fan, Shi, and Wang}]{xu2022sinnerf}
Xu, D.; Jiang, Y.; Wang, P.; Fan, Z.; Shi, H.; and Wang, Z. 2022.
\newblock Sinnerf: Training neural radiance fields on complex scenes from a
  single image.
\newblock In \emph{European Conference on Computer Vision}.

\bibitem[{Yao et~al.(2018)Yao, Luo, Li, Fang, and Quan}]{yao2018mvsnet}
Yao, Y.; Luo, Z.; Li, S.; Fang, T.; and Quan, L. 2018.
\newblock Mvsnet: Depth inference for unstructured multi-view stereo.
\newblock In \emph{Proceedings of the European conference on computer vision
  (ECCV)}.

\bibitem[{Yu et~al.(2021)Yu, Ye, Tancik, and Kanazawa}]{pixelnerf}
Yu, A.; Ye, V.; Tancik, M.; and Kanazawa, A. 2021.
\newblock {pixelNeRF}: Neural Radiance Fields from One or Few Images.
\newblock In \emph{Proceedings of the IEEE/CVF Conference on Computer Vision
  and Pattern Recognition (CVPR)}.

\bibitem[{Yuan et~al.(2022)Yuan, Sun, Lai, Ma, Jia, and Gao}]{yuan2022nerf}
Yuan, Y.-J.; Sun, Y.-T.; Lai, Y.-K.; Ma, Y.; Jia, R.; and Gao, L. 2022.
\newblock Nerf-editing: geometry editing of neural radiance fields.
\newblock In \emph{Proceedings of the IEEE/CVF Conference on Computer Vision
  and Pattern Recognition}.

\end{thebibliography}

\newpage

\appendix

\section{Results}
{

    \subsection{Multi-View Segmentation Results}
    {
        For multi-view segmentation, we use the user's stroke-based segmentation method. Similar to ISRF \cite{isrf}, we cluster the marked features using K-Means Clustering and then apply NNFM (Nearest Neighbour Feature Matching) between the cluster centres and the features predicted by our GSN model. This results in better segmentation than the naive average feature-matching approach.
        
        In the main paper, we show the results of the four scenes of the LLFF dataset \cite{mildenhall2019llff} and compare them against the state-of-the-art methods. In \cref{fig:supp_fourscenes}, we show our segmentation results on multiple views to demonstrate our multi-view segmentation capability. These results showcase our multi-view generalised segmentation capability, i.e. segmentation of multiple views of unseen data.

    }
}
\section{Validating Distillation}
{
    \begin{figure*}[!ht]
    \centering
    
    
    \rotatebox[origin=c]{90}{Chesstable}
    \begin{minipage}{0.32\linewidth}
    \centering
    \frame{\includegraphics[width=\textwidth]{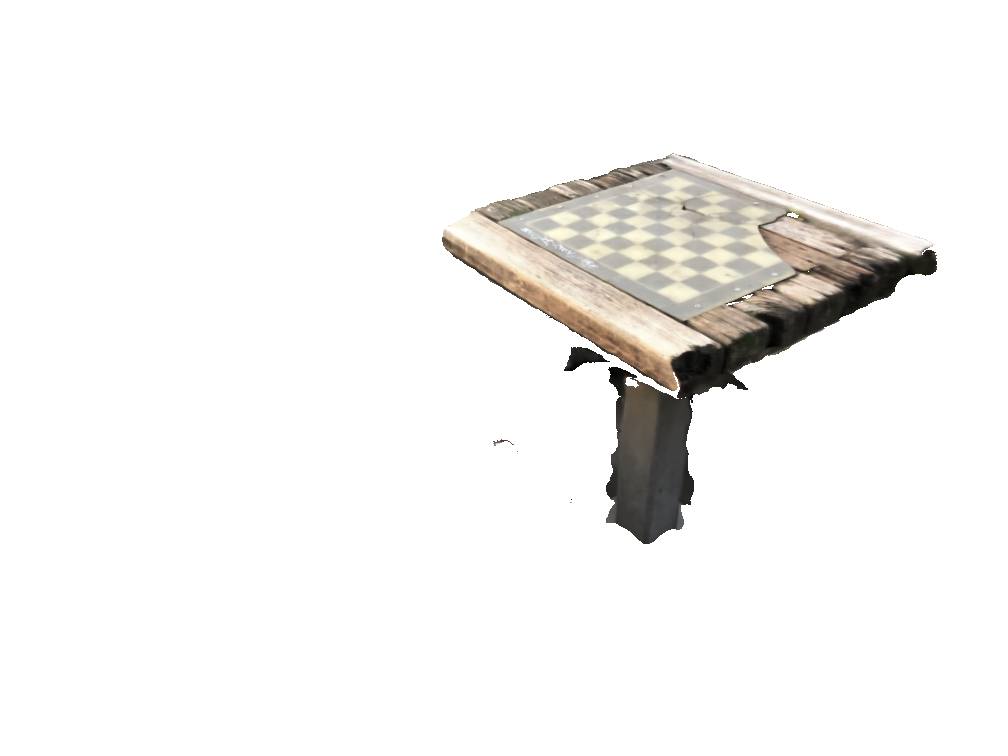}}
    \end{minipage}
    \begin{minipage}{0.32\linewidth}
        \centering
         \frame{\includegraphics[width=\textwidth]{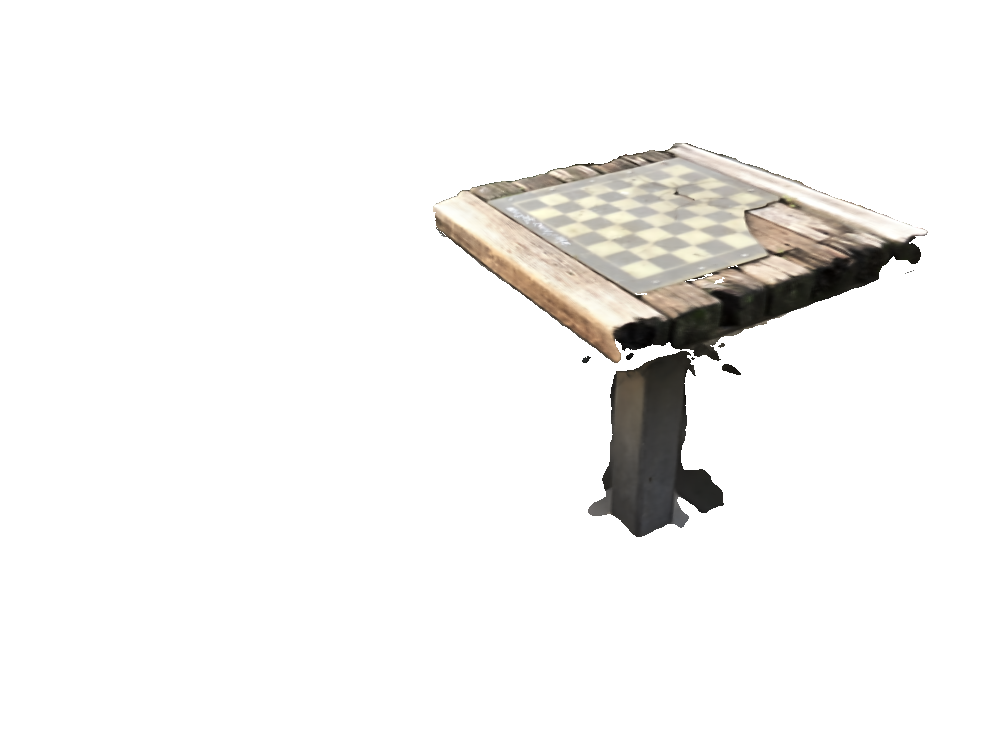}}
    \end{minipage}
    \begin{minipage}{0.32\linewidth}
        \centering
         \frame{\includegraphics[width=\textwidth]{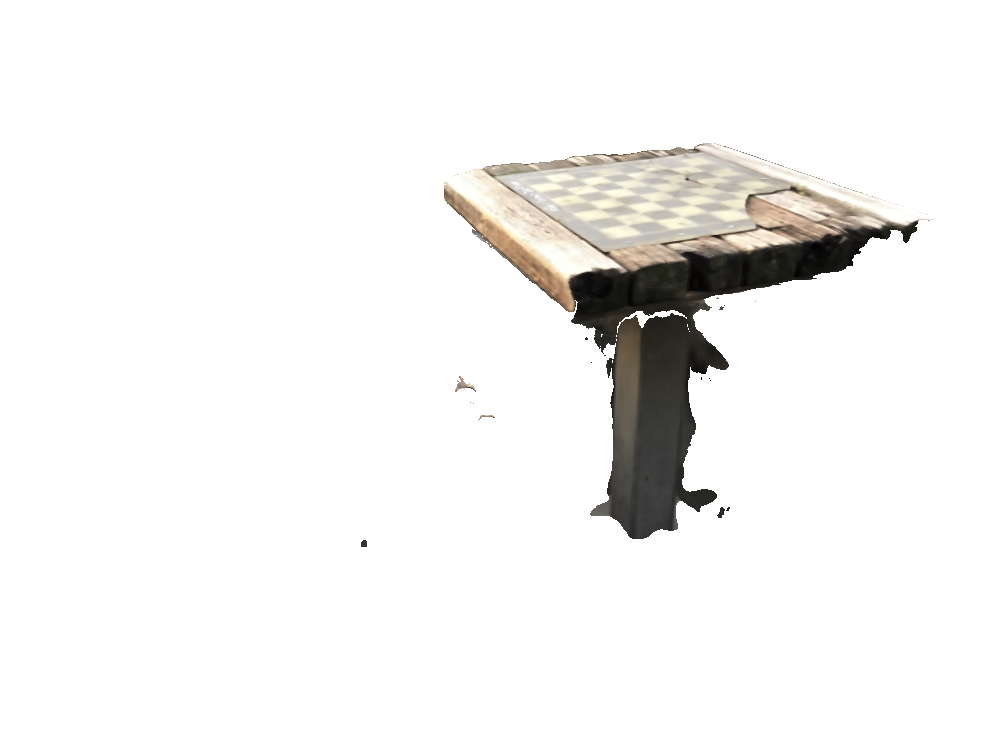}}
    \end{minipage}
    
    \rotatebox[origin=c]{90}{Colorfountain}
    \begin{minipage}{0.32\linewidth}
    \centering
    \frame{\includegraphics[width=\textwidth]{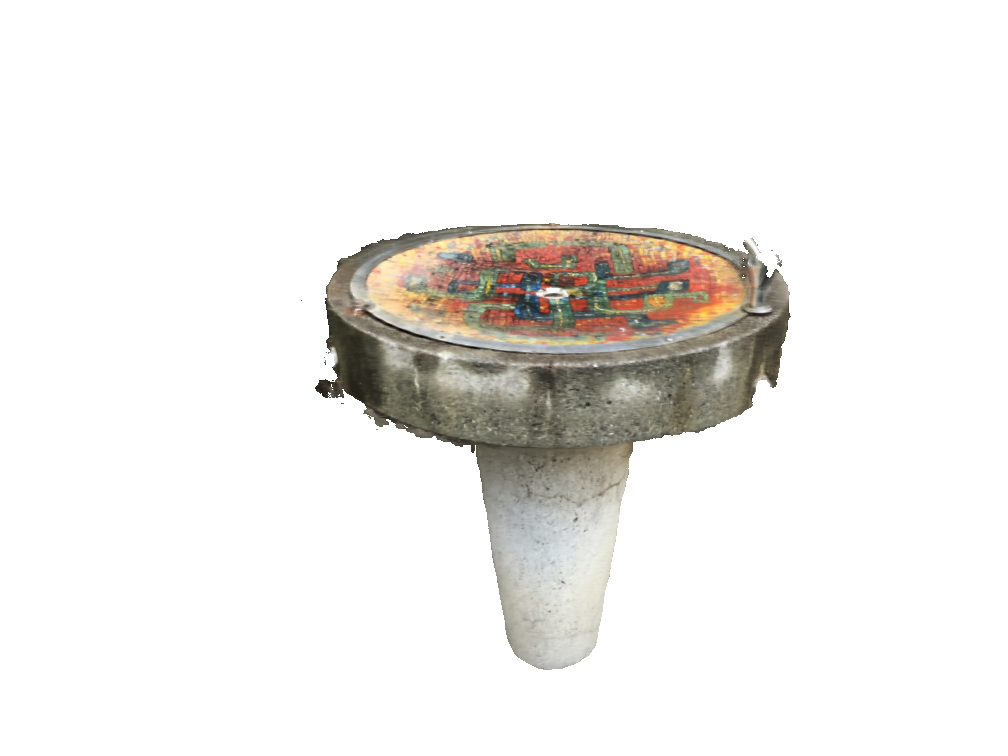}}
    \end{minipage}
    \begin{minipage}{0.32\linewidth}
        \centering
         \frame{\includegraphics[width=\textwidth]{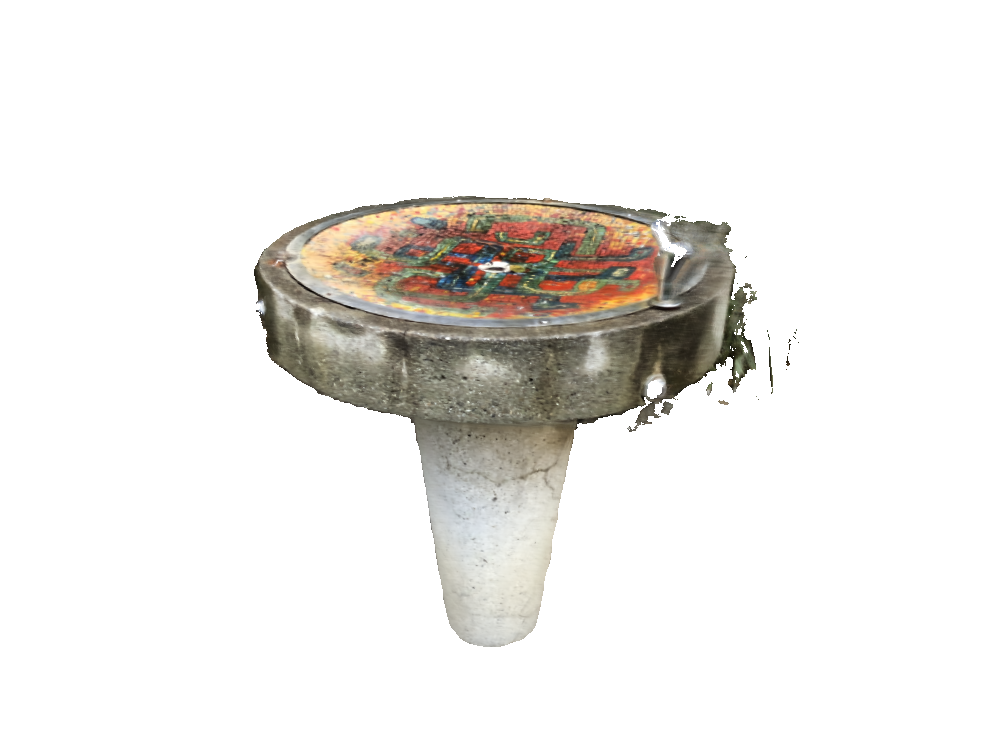}}
    \end{minipage}
    \begin{minipage}{0.32\linewidth}
        \centering
         \frame{\includegraphics[width=\textwidth]{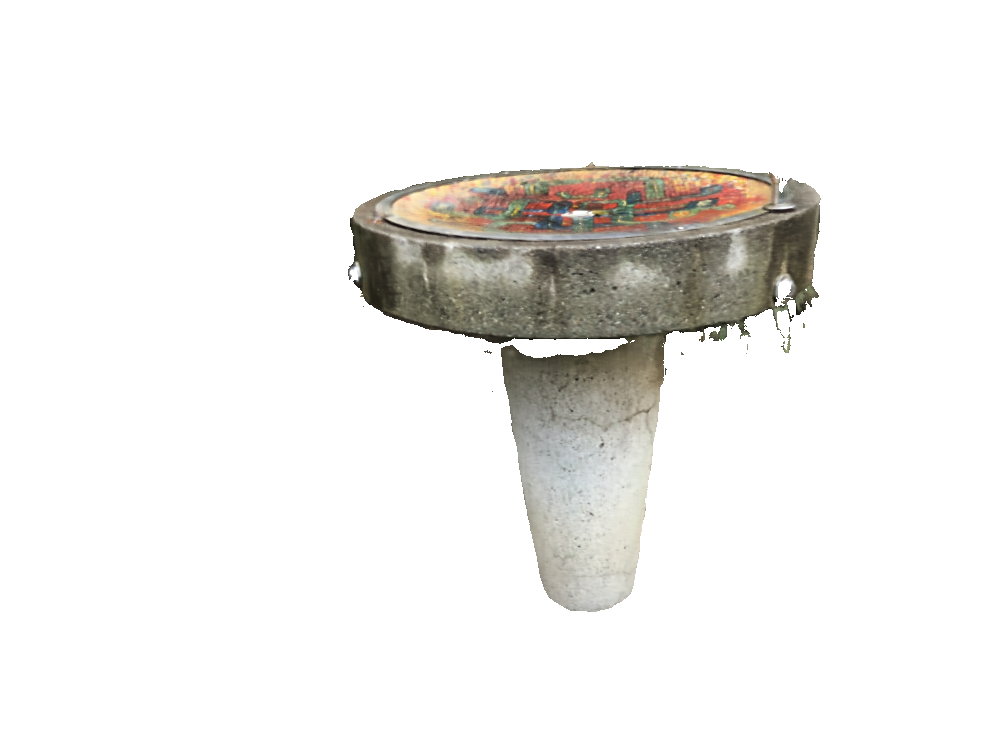}}
    \end{minipage}
    
    \rotatebox[origin=c]{90}{Shoerack}
    \begin{minipage}{0.32\linewidth}
    \centering
    \frame{\includegraphics[width=\textwidth]{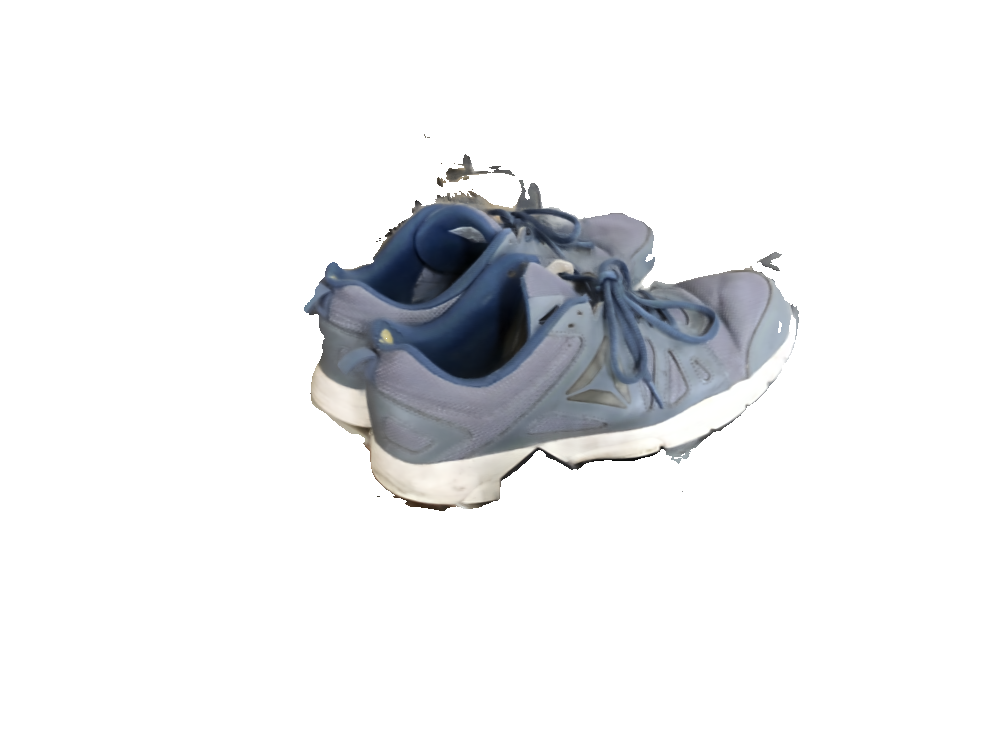}}
    \end{minipage}
    \begin{minipage}{0.32\linewidth}
        \centering
         \frame{\includegraphics[width=\textwidth]{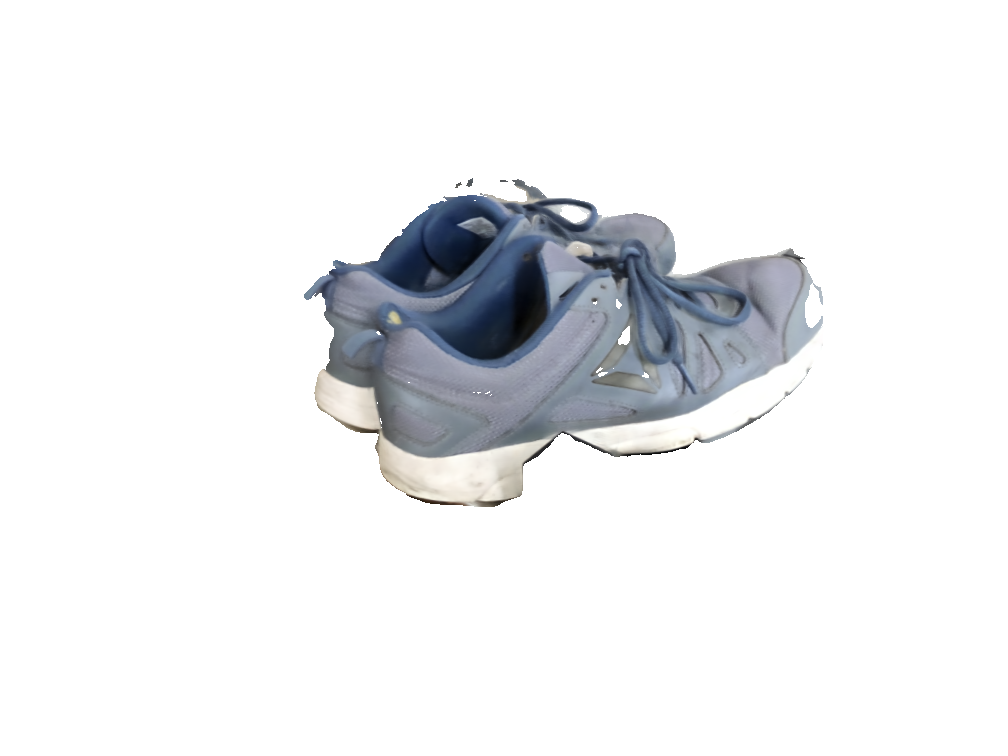}}
    \end{minipage}
    \begin{minipage}{0.32\linewidth}
        \centering
         \frame{\includegraphics[width=\textwidth]{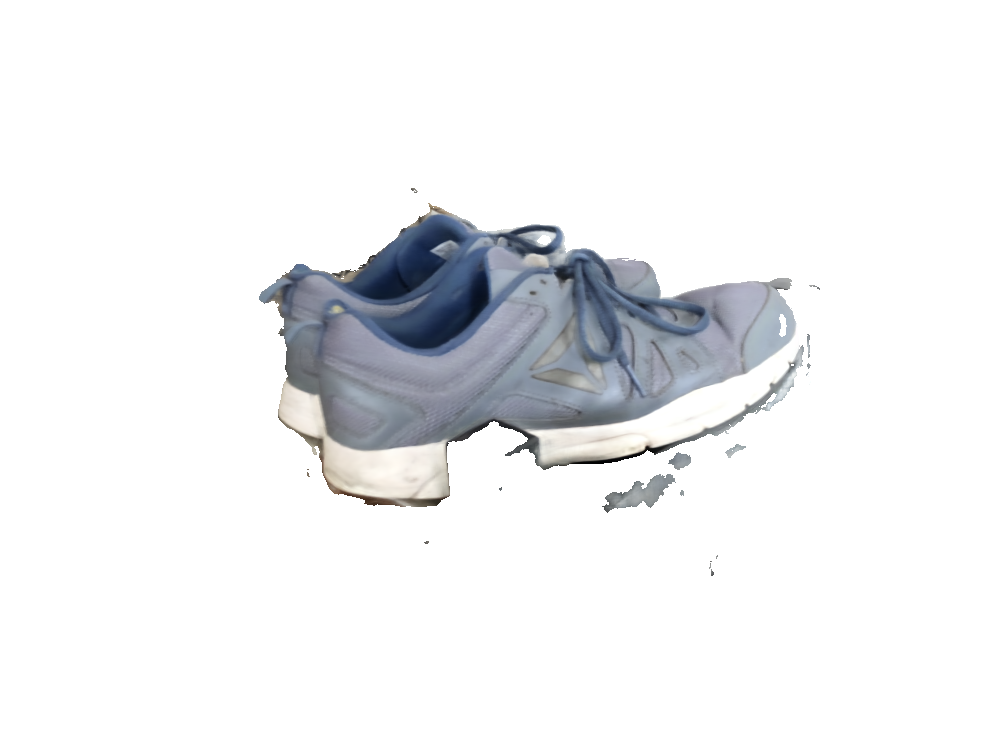}}
    \end{minipage}
    
    \rotatebox[origin=c]{90}{Stove}
    \begin{minipage}{0.32\linewidth}
    \centering
    \frame{\includegraphics[width=\textwidth]{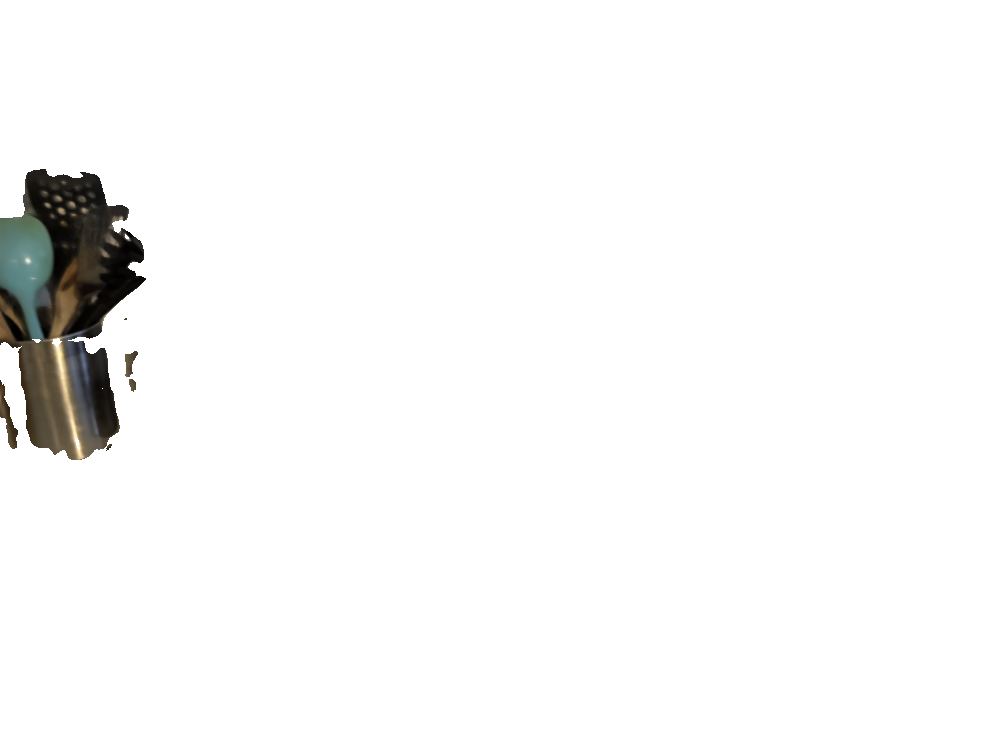}}
    \end{minipage}
    \begin{minipage}{0.32\linewidth}
        \centering
         \frame{\includegraphics[width=\textwidth]{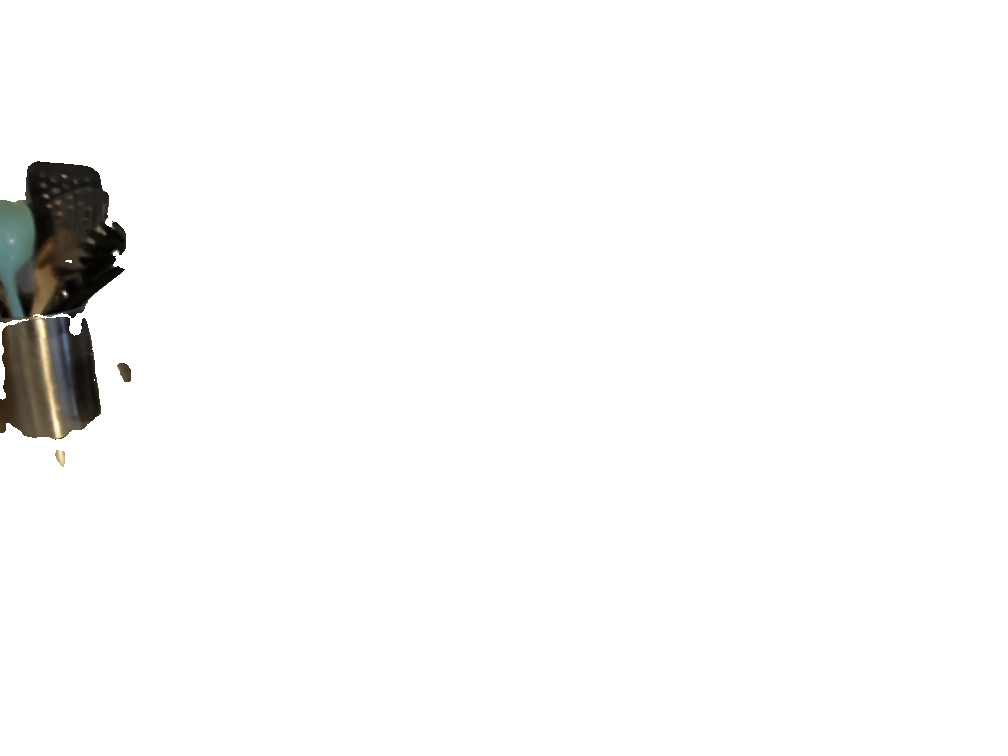}}
    \end{minipage}
    \begin{minipage}{0.32\linewidth}
        \centering
         \frame{\includegraphics[width=\textwidth]{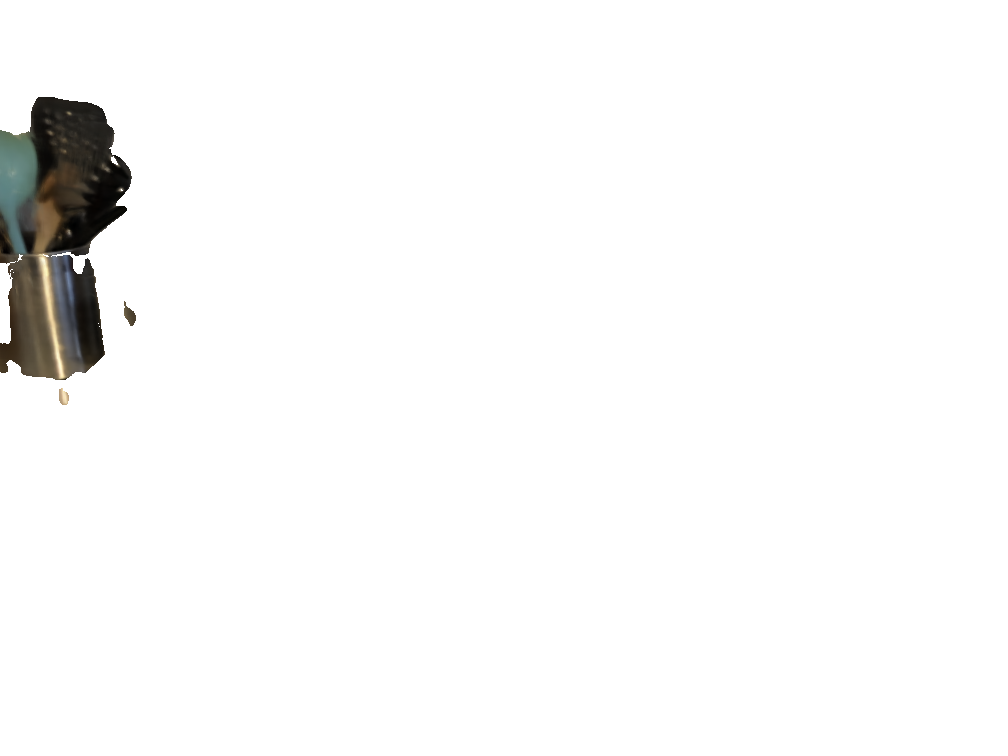}}
    \end{minipage}
    
    \caption{\label{fig:supp_fourscenes} {\em Comparison:} We show multi-view segmentation results on the four scenes in the main paper. We see similar results as shown in the main paper. The chesstable and the colorfountain scene have artefacts near the edges of the object, which are due to fewer views leading to wrong epipolar geometry. In the shoerack scene, the pole obstructing the shoe is consistently removed from all views. In the stove scene, the long black spoon is recovered correctly from all the different views.}
    
\end{figure*}

    \subsection{Student surpasses teacher}
    {
        In the main paper, we mentioned that the distilled features of the student model from GSN are far superior in granularity compared to other methods of obtaining the features. To emphasise this further, in \cref{fig:supp_flowerdino} and \cref{fig:supp_orchidsdino}, we provide results on two more scenes, \flower and \orchids LLFF dataset \cite{mildenhall2019llff}. In both figures, the fine granularity of our predicted features can be observed in the feature images (Row 2).
        
        In Row 3 of \cref{fig:supp_flowerdino} and \cref{fig:supp_orchidsdino}, we show the visualisation after applying K-Means clustering on the features predicted by their respective methods. This indicates that using DINO on the image generated by any radiance field causes very noisy and coarse features. In contrast, our method generates clean, noise-free features. This is because our method accounts for feature information from multiple views, unlike the other two methods.
        
        In \cref{fig:supp_viewconsis}, we show the view consistency in features between different views of the same scene by clustering the features (shown in different shades). We see that across views, our model can generate consistent features, whereas the other methods generate view-dependent coarse features which don't perform well for downstream tasks (demonstrated using clustering).

        Our student GSN model distils DINO features from a pre-trained 2D DINO transformer, creating a semantic feature field along with the Radiance Fields. We observed that the student(GSN) model surpasses the teacher model post-distillation in generating semantic labels alongside radiance.

    }
    \begin{figure*}[!b]
    
    \begin{minipage}{0.24\linewidth}
        \centering
        \begin{tikzpicture}
            \node[anchor=south west,inner sep=0] (image) at (0,0) {\includegraphics[width=\textwidth]{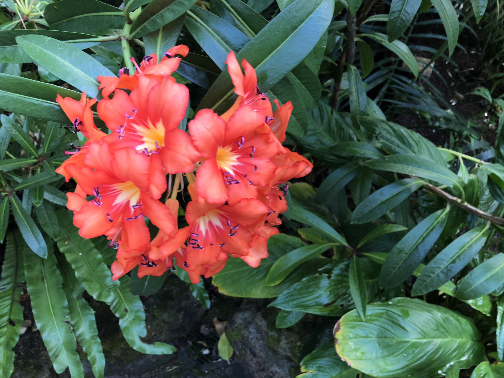}};
        \end{tikzpicture}
    \end{minipage}
    \begin{minipage}{0.24\linewidth}
        \centering
        \begin{tikzpicture}
            \node[anchor=south west,inner sep=0] (image) at (0,0) {\includegraphics[width=\textwidth]{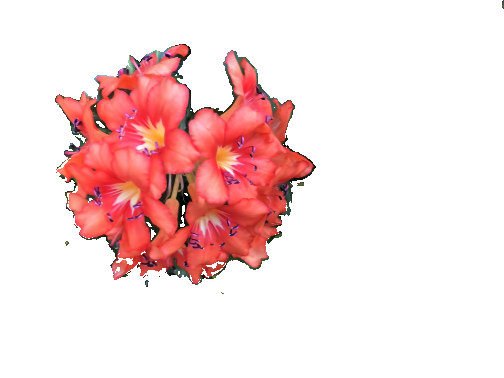}};
        \end{tikzpicture}
    \end{minipage}
    \begin{minipage}{0.24\linewidth}
        \centering
        \begin{tikzpicture}
            \node[anchor=south west,inner sep=0] (image) at (0,0) {\includegraphics[width=\textwidth]{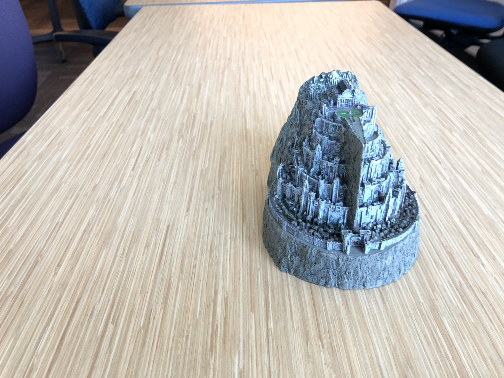}};
        \end{tikzpicture}
    \end{minipage}
    \begin{minipage}{0.24\linewidth}
        \centering
        \begin{tikzpicture}
            \node[anchor=south west,inner sep=0] (image) at (0,0) {\includegraphics[width=\textwidth]{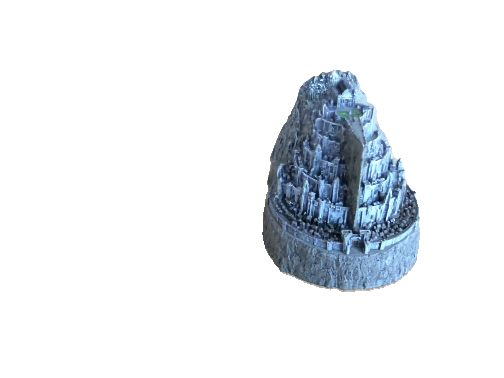}};
        \end{tikzpicture}
    \end{minipage}

    \begin{minipage}{0.24\linewidth}
        \centering
        \begin{tikzpicture}
            \node[anchor=south west,inner sep=0] (image) at (0,0) {\includegraphics[width=\textwidth]{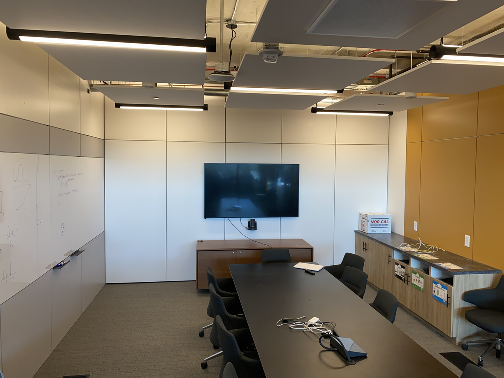}};
        \end{tikzpicture}
    \end{minipage}
    \begin{minipage}{0.24\linewidth}
        \centering
        \begin{tikzpicture}
            \node[anchor=south west,inner sep=0] (image) at (0,0) {\includegraphics[width=\textwidth]{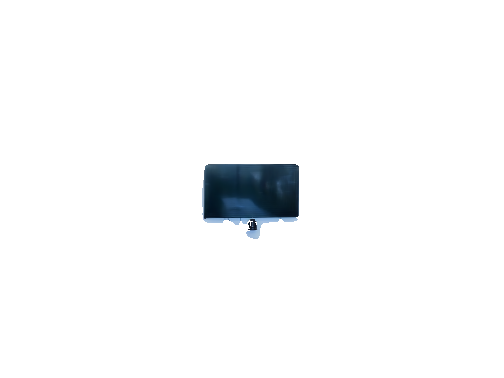}};
        \end{tikzpicture}
    \end{minipage}
    \begin{minipage}{0.24\linewidth}
        \centering
        \begin{tikzpicture}
            \node[anchor=south west,inner sep=0] (image) at (0,0) {\includegraphics[width=\textwidth]{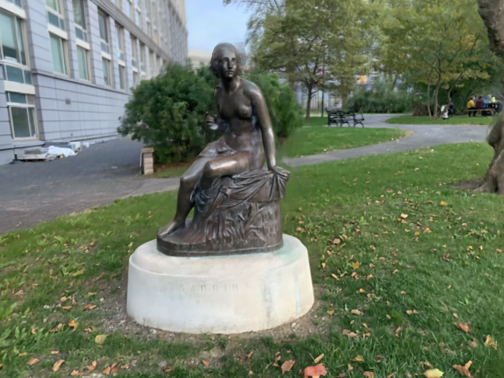}};
        \end{tikzpicture}
    \end{minipage}
    \begin{minipage}{0.24\linewidth}
        \centering
        \begin{tikzpicture}
            \node[anchor=south west,inner sep=0] (image) at (0,0) {\includegraphics[width=\textwidth]{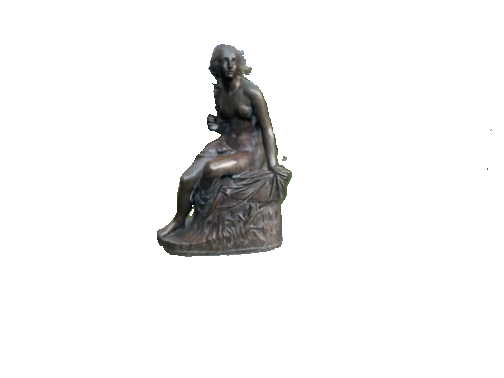}};
        \end{tikzpicture}
    \end{minipage}

    \begin{minipage}{0.24\linewidth}
        \centering
        \begin{tikzpicture}
            \node[anchor=south west,inner sep=0] (image) at (0,0) {\includegraphics[width=\textwidth]{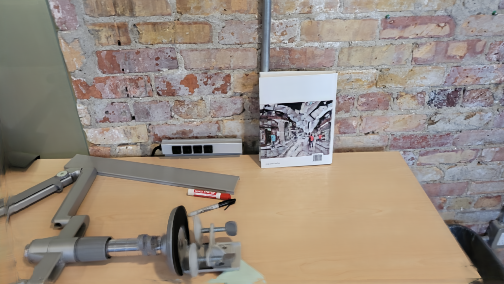}};
        \end{tikzpicture}
    \end{minipage}
    \begin{minipage}{0.24\linewidth}
        \centering
        \begin{tikzpicture}
            \node[anchor=south west,inner sep=0] (image) at (0,0) {\includegraphics[width=\textwidth]{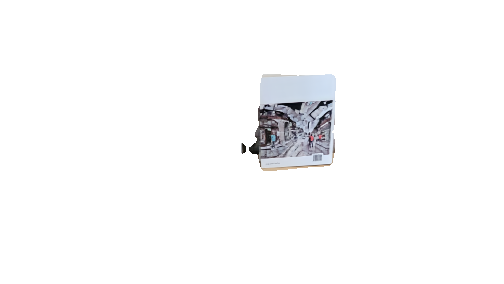}};
        \end{tikzpicture}
    \end{minipage}
    \begin{minipage}{0.24\linewidth}
        \centering
        \begin{tikzpicture}
            \node[anchor=south west,inner sep=0] (image) at (0,0) {\includegraphics[width=\textwidth]{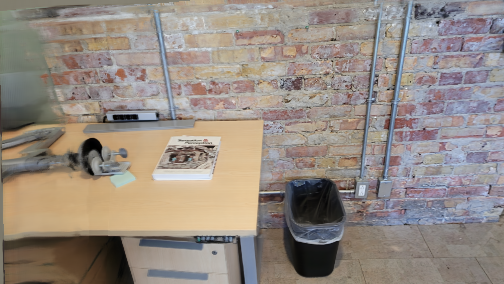}};
        \end{tikzpicture}
    \end{minipage}
    \begin{minipage}{0.24\linewidth}
        \centering
        \begin{tikzpicture}
            \node[anchor=south west,inner sep=0] (image) at (0,0) {\includegraphics[width=\textwidth]{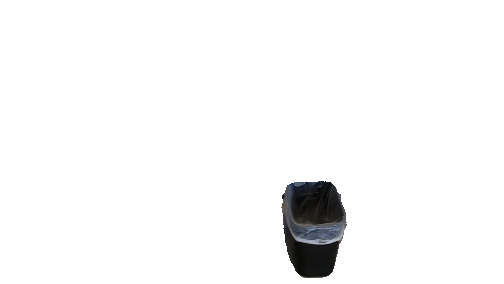}};
        \end{tikzpicture}
    \end{minipage}

    \begin{minipage}{0.24\linewidth}
        \centering
        \begin{tikzpicture}
            \node[anchor=south west,inner sep=0] (image) at (0,0) {\includegraphics[width=\textwidth]{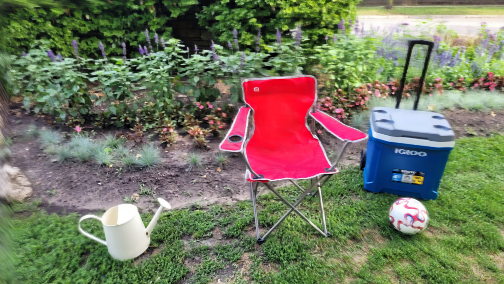}};
        \end{tikzpicture}
    \end{minipage}
    \begin{minipage}{0.24\linewidth}
        \centering
        \begin{tikzpicture}
            \node[anchor=south west,inner sep=0] (image) at (0,0) {\includegraphics[width=\textwidth]{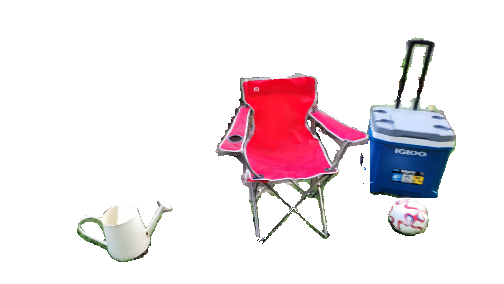}};
        \end{tikzpicture}
    \end{minipage}
    \begin{minipage}{0.24\linewidth}
        \centering
        \begin{tikzpicture}
            \node[anchor=south west,inner sep=0] (image) at (0,0) {\includegraphics[width=\textwidth]{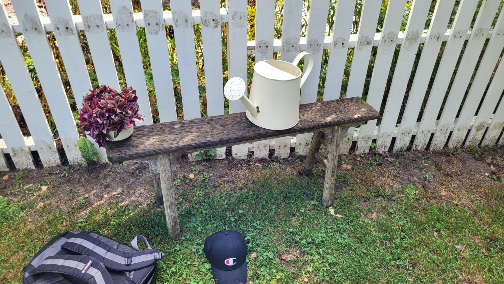}};
        \end{tikzpicture}
    \end{minipage}
    \begin{minipage}{0.24\linewidth}
        \centering
        \begin{tikzpicture}
            \node[anchor=south west,inner sep=0] (image) at (0,0) {\includegraphics[width=\textwidth]{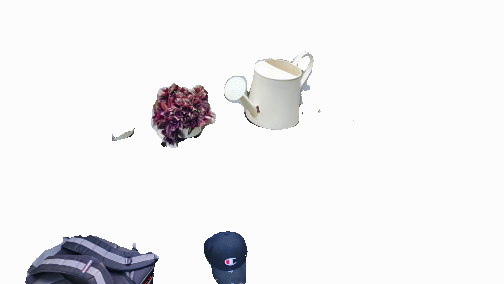}};
        \end{tikzpicture}
    \end{minipage}

    \caption{\label{fig:supp_moreresults}  We show segmentation results on more scenes from the NeRF-LLFF\cite{nerf} and Spin-NeRF\cite{spinnerf} dataset. The first 3 rows shows single object segmentation using our DINO integrated model. The last row shows multi-object segmentation on the 2 scenes from the Spin-NeRF dataset. Note: For the multi-object segmentation, we segment each object individually and combine them into one image for visualisation purpose. }

\end{figure*}
    \input{src/figures/supp_betterthandino}
    \begin{figure*}[!ht]
    \begin{minipage}{0.24\linewidth}
        \centering
        Reference
    \end{minipage}
    \begin{minipage}{0.24\linewidth}
        \centering
        Original 
    \end{minipage}
    \begin{minipage}{0.24\linewidth}
        \centering
        GNT
    \end{minipage}
    \begin{minipage}{0.24\linewidth}
        \centering
        GSN (Ours)
    \end{minipage}
    
    \rotatebox[origin=c]{90}{View 1}
    \begin{minipage}{0.24\linewidth}
        \centering
        \begin{tikzpicture}
            \node[anchor=south west,inner sep=0] (image) at (0,0) {\includegraphics[width=\textwidth]{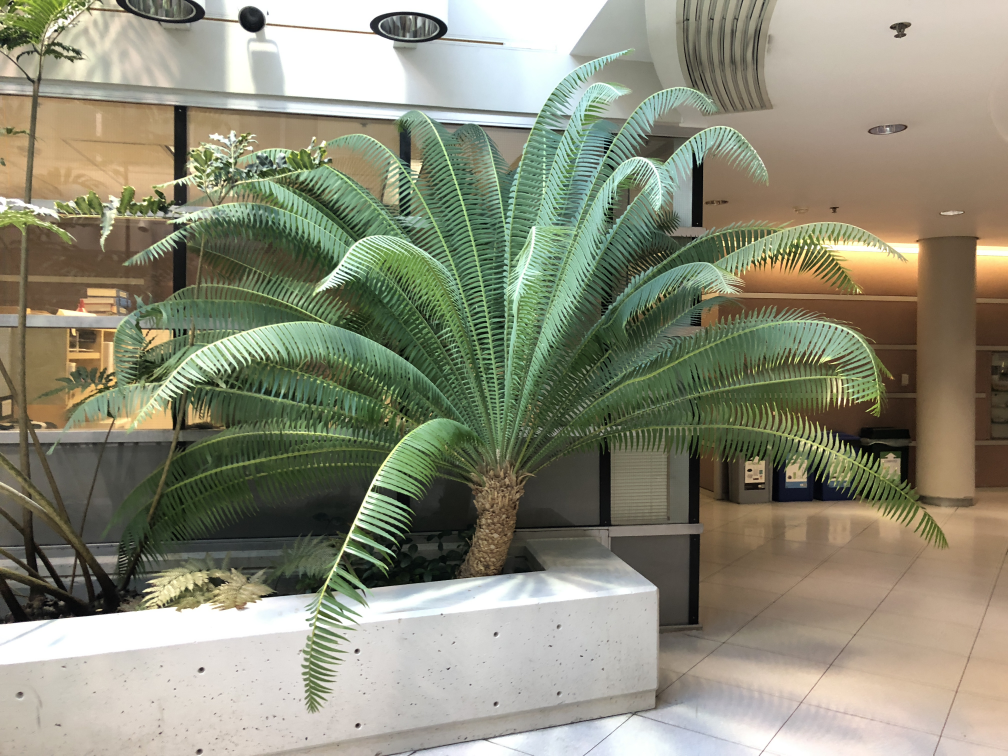}};
        \end{tikzpicture}
    \end{minipage}
    \begin{minipage}{0.24\linewidth}
        \centering
        \begin{tikzpicture}
            \node[anchor=south west,inner sep=0] (image) at (0,0) {\includegraphics[width=\textwidth]{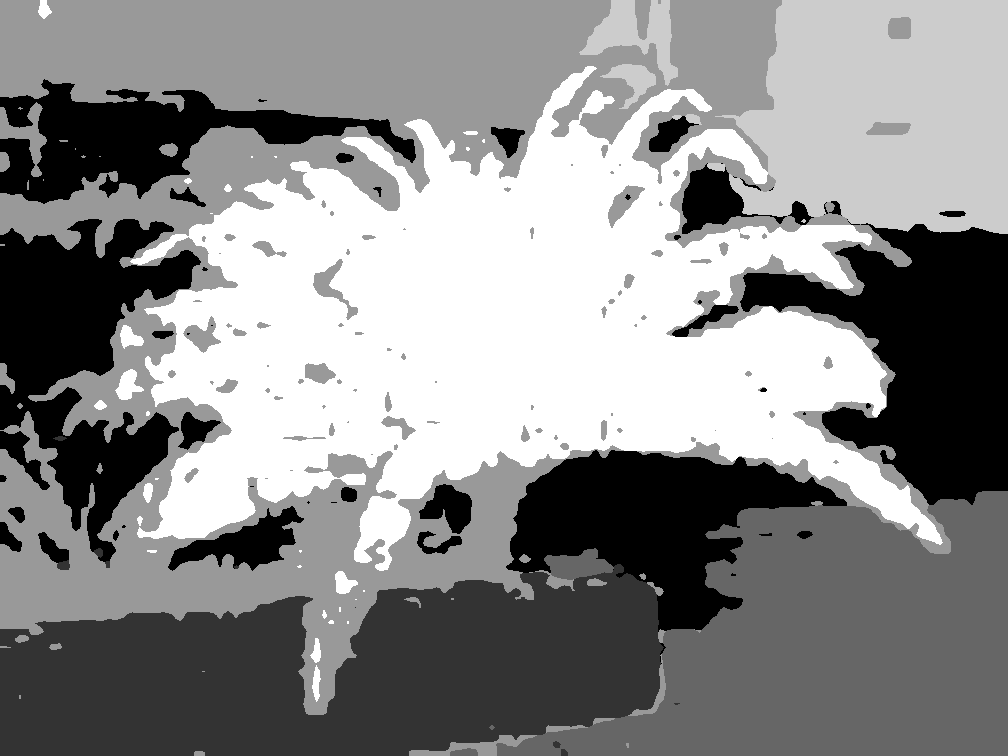}};
            \begin{scope}[x={(image.south east)},y={(image.north west)}]
                \draw[cyan,thick] (0.1,0.99) rectangle (0.6, 0.8);
                \draw[magenta,thick] (0.4, 0.45) rectangle (0.6, 0.2);
            \end{scope}
        \end{tikzpicture}
    \end{minipage}
    \begin{minipage}{0.24\linewidth}
        \centering
        \begin{tikzpicture}
            \node[anchor=south west,inner sep=0] (image) at (0,0) {\includegraphics[width=\textwidth]{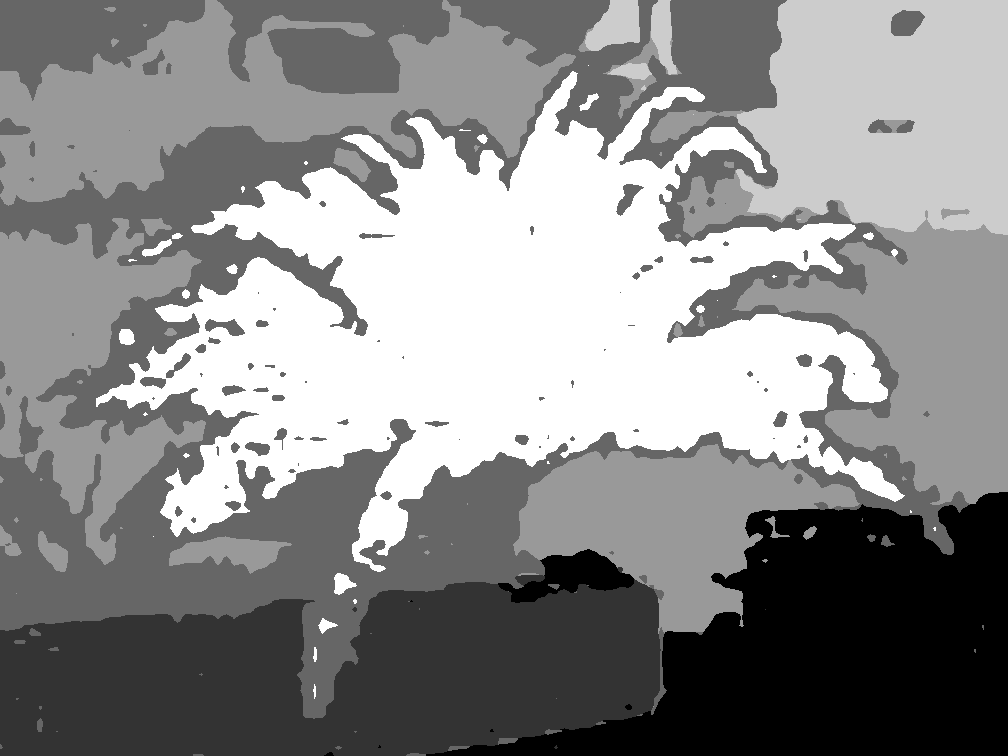}};
            \begin{scope}[x={(image.south east)},y={(image.north west)}]
                \draw[cyan,thick] (0.1,0.99) rectangle (0.6, 0.8);
                \draw[magenta,thick] (0.5, 0.3) rectangle (0.9, 0.01);
            \end{scope}
        \end{tikzpicture}
    \end{minipage}
    \begin{minipage}{0.24\linewidth}
        \centering
        \begin{tikzpicture}
            \node[anchor=south west,inner sep=0] (image) at (0,0) {\includegraphics[width=\textwidth]{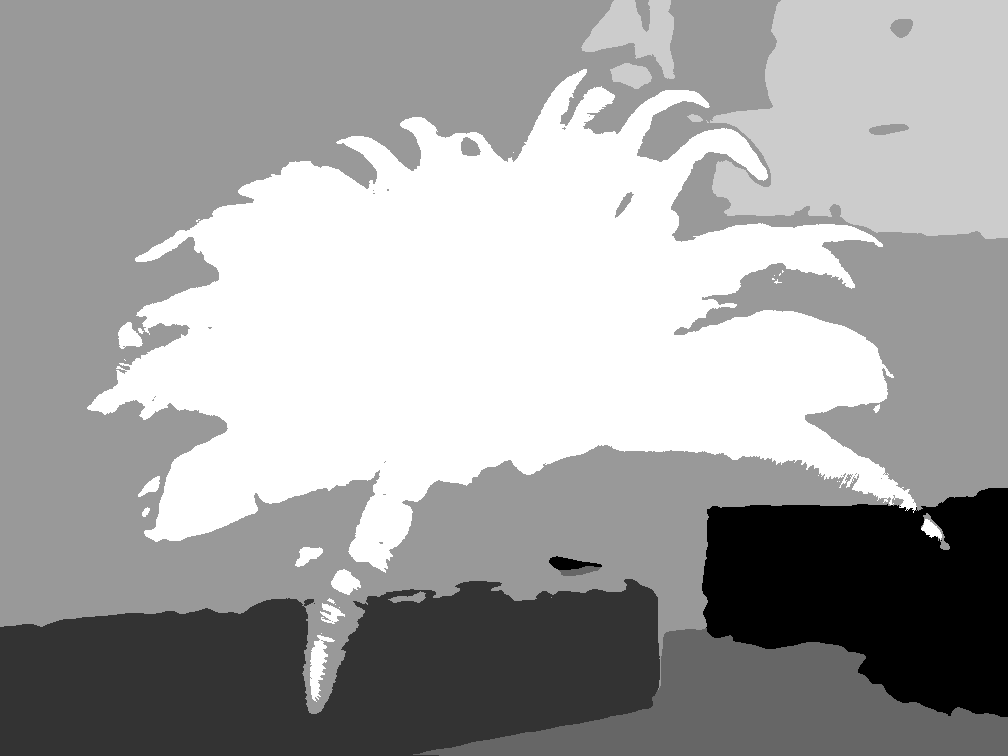}};
            \begin{scope}[x={(image.south east)},y={(image.north west)}]
                \draw[cyan,thick] (0.1,0.99) rectangle (0.6, 0.8);
                \draw[magenta,thick] (0.45, 0.4) rectangle (0.7, 0.15);
            \end{scope}
        \end{tikzpicture}
    \end{minipage}

    \rotatebox[origin=c]{90}{View 2}
    \begin{minipage}{0.24\linewidth}
        \centering
        \begin{tikzpicture}
            \node[anchor=south west,inner sep=0] (image) at (0,0) {\includegraphics[width=\textwidth]{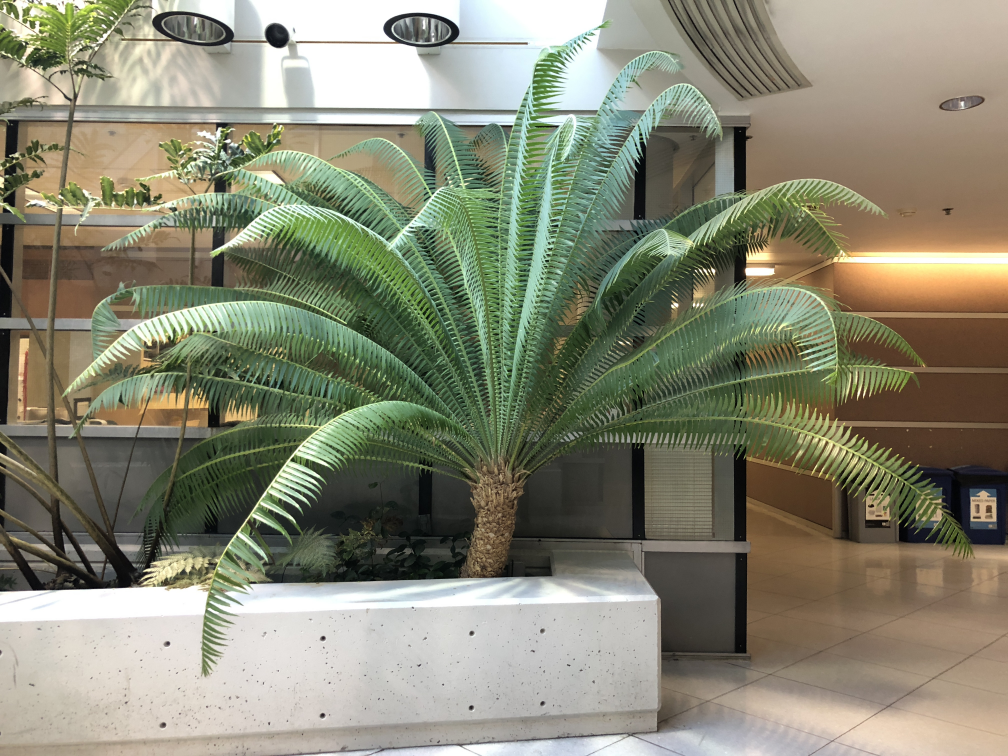}};
        \end{tikzpicture}
    \end{minipage}
    \begin{minipage}{0.24\linewidth}
        \centering
        \begin{tikzpicture}
            \node[anchor=south west,inner sep=0] (image) at (0,0) {\includegraphics[width=\textwidth]{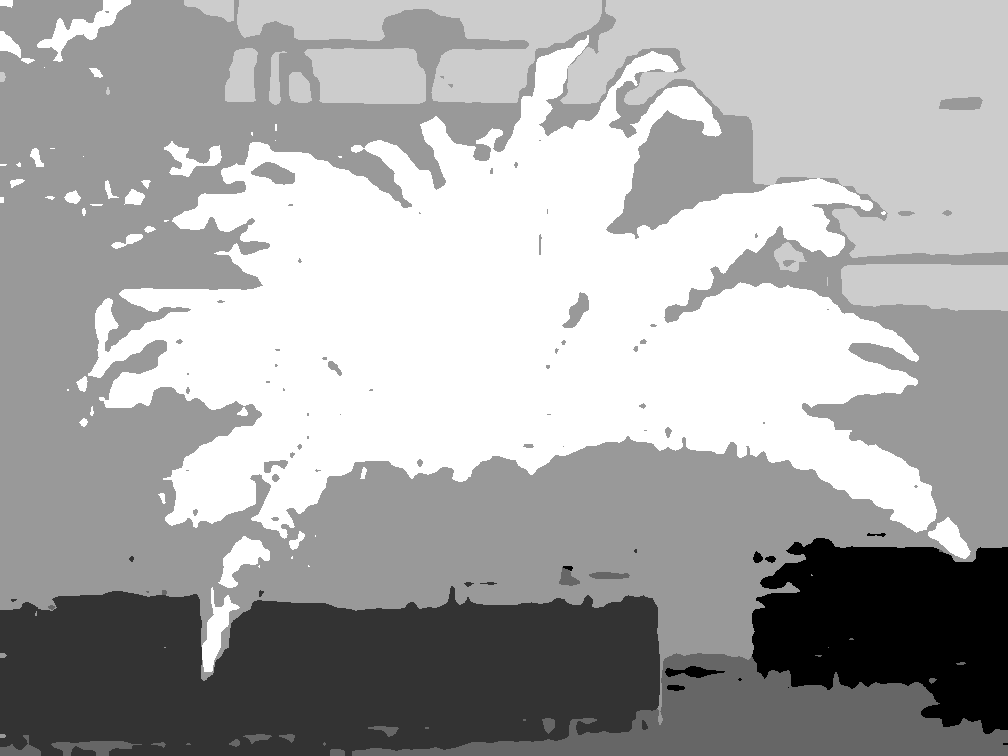}};
            \begin{scope}[x={(image.south east)},y={(image.north west)}]
                \draw[cyan,thick] (0.1,0.99) rectangle (0.6, 0.8);
                \draw[magenta,thick] (0.4, 0.45) rectangle (0.6, 0.2);
            \end{scope}
        \end{tikzpicture}
    \end{minipage}
    \begin{minipage}{0.24\linewidth}
        \centering
        \begin{tikzpicture}
            \node[anchor=south west,inner sep=0] (image) at (0,0) {\includegraphics[width=\textwidth]{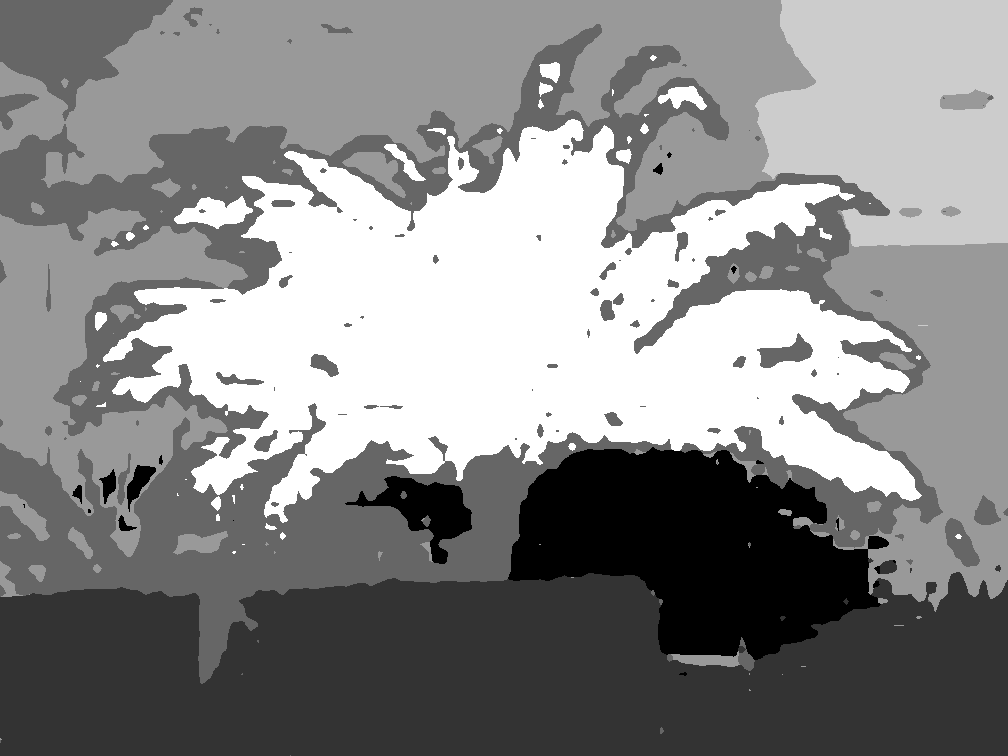}};
            \begin{scope}[x={(image.south east)},y={(image.north west)}]
                \draw[cyan,thick] (0.1,0.99) rectangle (0.6, 0.8);
                \draw[magenta,thick] (0.5, 0.3) rectangle (0.9, 0.01);
            \end{scope}
        \end{tikzpicture}
    \end{minipage}
    \begin{minipage}{0.24\linewidth}
        \centering
        \begin{tikzpicture}
            \node[anchor=south west,inner sep=0] (image) at (0,0) {\includegraphics[width=\textwidth]{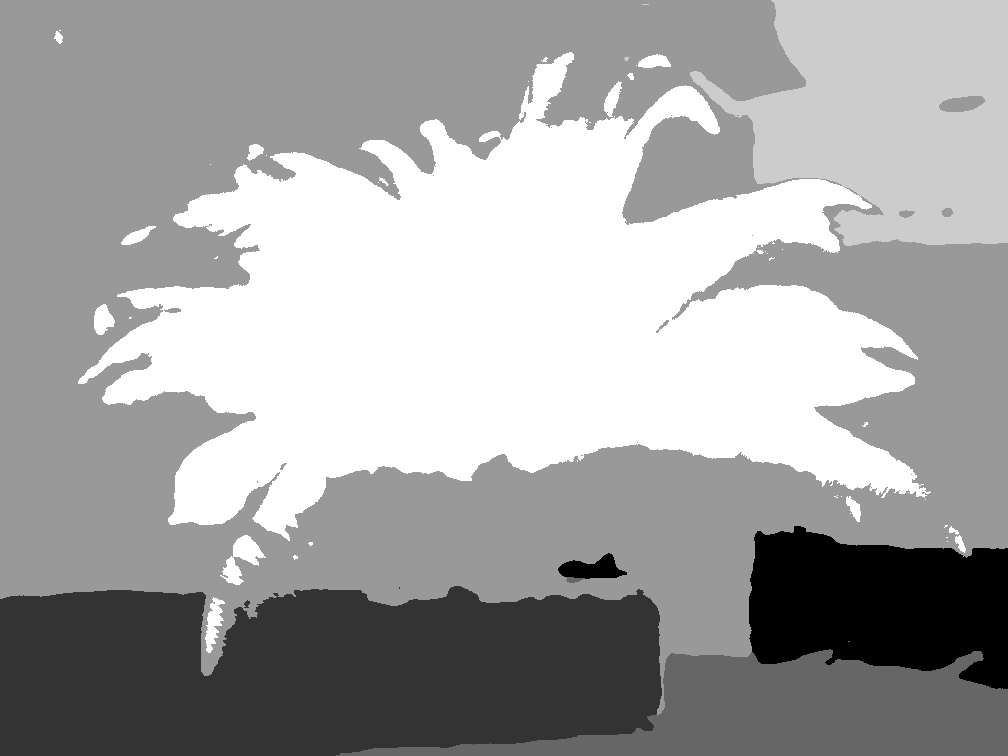}};
            \begin{scope}[x={(image.south east)},y={(image.north west)}]
                \draw[cyan,thick] (0.1,0.99) rectangle (0.6, 0.8);
                \draw[magenta,thick] (0.45, 0.4) rectangle (0.7, 0.15);
            \end{scope}
        \end{tikzpicture}
    \end{minipage}

    \caption{\label{fig:supp_viewconsis} In this figure, we show the multi-view consistency of our model. We render the feature images using our method for two different views of the \fern scene in LFF dataset \cite{mildenhall2019llff}. These features are clustered into $6$ clusters using K-Means Clustering. In the case of GNT \cite{varma2022attention}, we observe that the ceiling and floor in views one and two have been clustered very differently. In the case of the original image, the stem of the fern plant has been segmented separately in view $1$, but in view $2$, it has blended with the background, leading to view inconsistency. Our method generates multi-view consistent features which give view-consistent clusters as highlighted in the boxes.}

\end{figure*}

    \subsection{Other Semantic Fields}
    {
        In this section, we elaborate on how we integrate other semantic fields like DINOv2 \cite{dinov2}, SAM \cite{sam}, and CLIP \cite{clip} into GSN. Please refer to \cref{fig:supp_othersemantics} alongside.

        \textbf{DINOv2}: DINOv2 performs excellent part segmentation, even across widely different images. We replicate their experimental setting using our improved GSN features. We first predict the features of a view and then run PCA on them to reduce them to 3 dimensions. The first dimension is thresholded (common across images) to remove the background. We again apply PCA on the foreground features and then visualise them in \cref{fig:supp_othersemantics}. For our case, we use DINOv2 ViT-L14 which gives $1024$ dimensional features. We PCA-reduce them to $64$ before performing distillation.
        
        \textbf{CLIP}: For text-based segmentation, we integrate OpenCLIP \cite{cherti2023reproducible} ViTB/16 model trained on the LAION-2B dataset similar to LeRF \cite{lerf}. LeRF solves the problem of text-based segmentation by integrating CLIP in a scale-wise manner of varying patch size to capture features at varying scales. This allows them to capture both global and local features. For the integration of CLIP into GSN, we extract features at a single scale with patch size $96 \times 96$ and stride $48$ in a sliding window fashion. This enables us to capture local features at a patch level instead of global features for the entire image. We aim to show that it is possible to integrate any set of features into GSN without much hassle.

        OpenCLIP text encoder outputs a 512 dimensional feature vector. For our purpose, we reduce it to 64 dimensions using PCA so that it can be compared against the features generated by our model. For this particular purpose, the PCA matrix for each scene is saved on disk.
        
        \textbf{SAM}: We integrate SAM features \cite{sam} and perform segmentation using them. We provide clicks on the image as a prompt to the model for segmenting a particular object. Similar to previous experiments, we extract the SAM features for each image of the scene and save them after running a PCA to reduce them to 64 dimensions. We used SAM ViT-B, which generates 256-dimensional features. We perform a 64-dimensional PCA on it and then integrate it into our GSN model. The SAM decoder that performs the segmentation requires 256-dimensional feature vectors along with the user clicks. We apply the inverse of the saved PCA matrix on the features before feeding them to the SAM decoder. This results in a slight loss of information, but our experiments are limited to 64 dimensional features in GSN due to computation limits.

    }
    \begin{figure*}[!ht]
    \begin{minipage}{0.32\linewidth}
        \centering
        CLIP
    \end{minipage}
    \begin{minipage}{0.32\linewidth}
        \centering
        DINO-v2
    \end{minipage}
    \begin{minipage}{0.32\linewidth}
        \centering
        SAM
    \end{minipage}
    
    \rotatebox[origin=c]{90}{Flower}
    \begin{minipage}{0.32\linewidth}
        \centering
        \begin{tikzpicture}
            \node[anchor=south west,inner sep=0] (image) at (0,0) {\includegraphics[width=\textwidth]{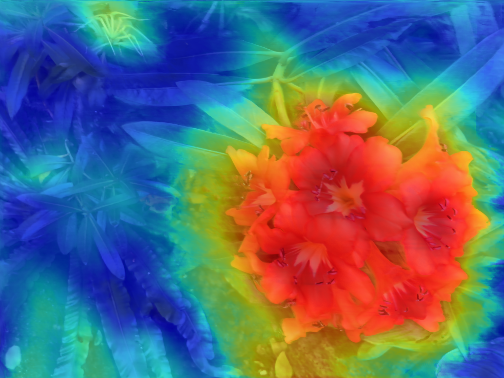}};
        \end{tikzpicture}
    \end{minipage}
    \begin{minipage}{0.32\linewidth}
        \centering
        \begin{tikzpicture}
            \node[anchor=south west,inner sep=0] (image) at (0,0) {\includegraphics[width=\textwidth]{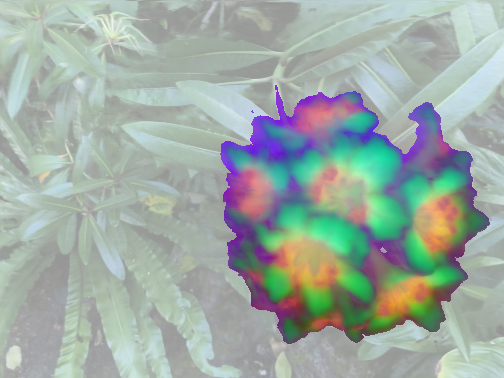}};
        \end{tikzpicture}
    \end{minipage}
    \begin{minipage}{0.32\linewidth}
        \centering
        \begin{tikzpicture}
            \node[anchor=south west,inner sep=0] (image) at (0,0) {\includegraphics[width=\textwidth]{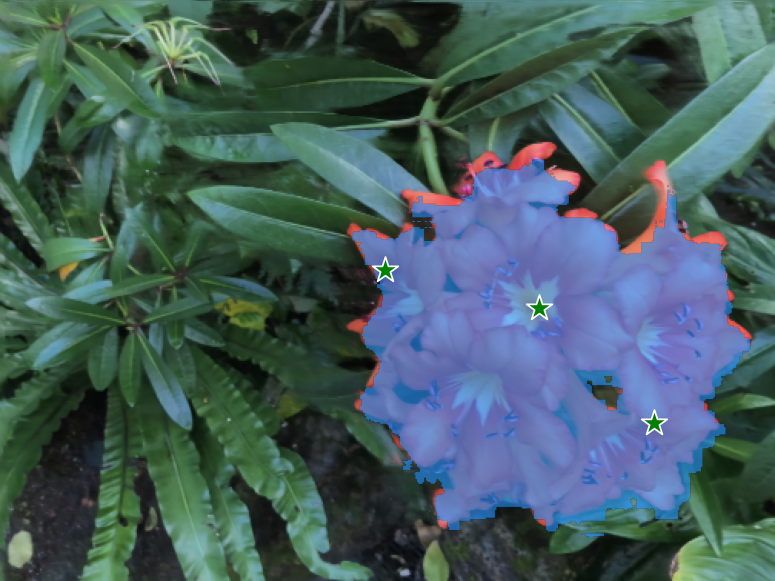}};
        \end{tikzpicture}
    \end{minipage}

    \rotatebox[origin=c]{90}{Fortress}
    \begin{minipage}{0.32\linewidth}
        \centering
        \begin{tikzpicture}
            \node[anchor=south west,inner sep=0] (image) at (0,0) {\includegraphics[width=\textwidth]{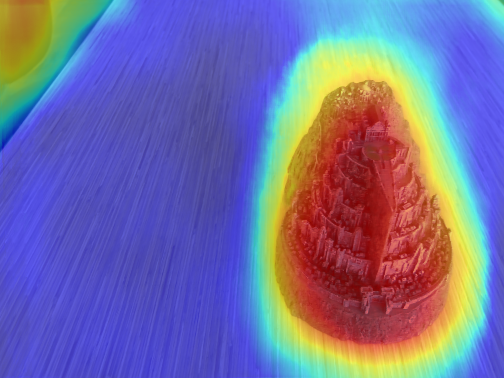}};
        \end{tikzpicture}
    \end{minipage}
    \begin{minipage}{0.32\linewidth}
        \centering
        \begin{tikzpicture}
            \node[anchor=south west,inner sep=0] (image) at (0,0) {\includegraphics[width=\textwidth]{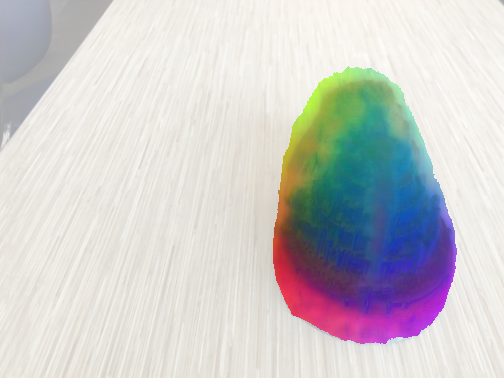}};
        \end{tikzpicture}
    \end{minipage}
    \begin{minipage}{0.32\linewidth}
        \centering
        \begin{tikzpicture}
            \node[anchor=south west,inner sep=0] (image) at (0,0) {\includegraphics[width=\textwidth]{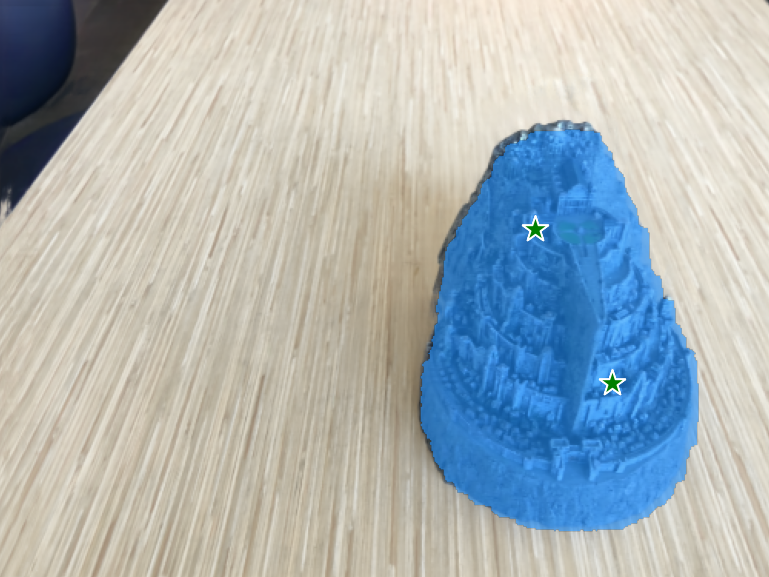}};
        \end{tikzpicture}
    \end{minipage}

    \rotatebox[origin=c]{90}{Trex}
    \begin{minipage}{0.32\linewidth}
        \centering
        \begin{tikzpicture}
            \node[anchor=south west,inner sep=0] (image) at (0,0) {\includegraphics[width=\textwidth]{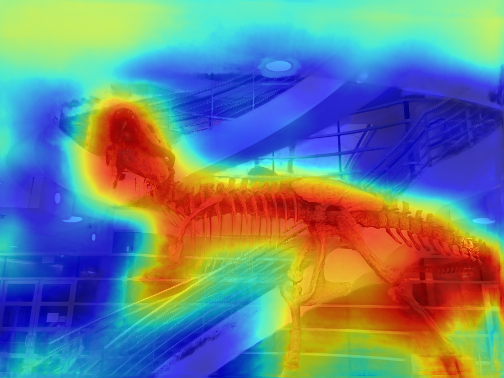}};
        \end{tikzpicture}
    \end{minipage}
    \begin{minipage}{0.32\linewidth}
        \centering
        \begin{tikzpicture}
            \node[anchor=south west,inner sep=0] (image) at (0,0) {\includegraphics[width=\textwidth]{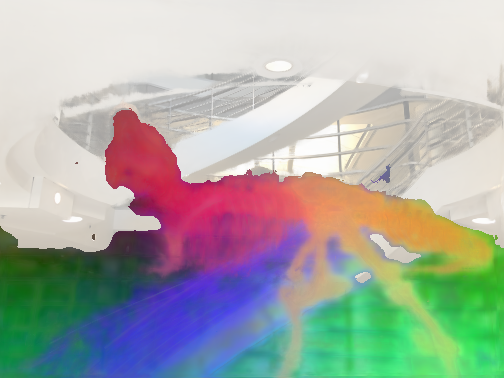}};
        \end{tikzpicture}
    \end{minipage}
    \begin{minipage}{0.32\linewidth}
        \centering
        \begin{tikzpicture}
            \node[anchor=south west,inner sep=0] (image) at (0,0) {\includegraphics[width=\textwidth]{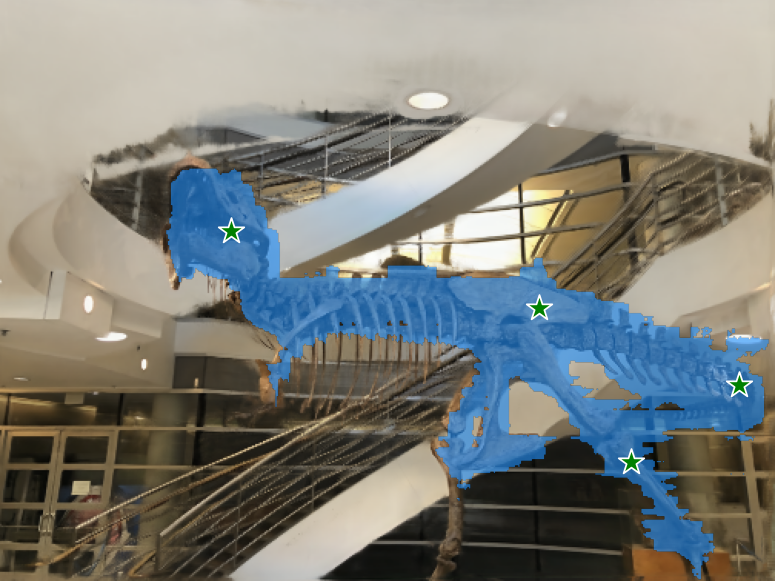}};
        \end{tikzpicture}
    \end{minipage}

    \caption{\label{fig:supp_othersemantics} This figure shows results of other semantic fields like DINO-v2, CLIP and SAM integrated into our method on three different scenes of the LLFF dataset \cite{mildenhall2019llff}. In the case of text-based segmentation using CLIP, we provide prompts such as {\em a flower}, {\em a fortress}, and {\em a fossil of dinosaur} to the following scenes and obtain the heatmap. Red and blue parts represent parts close to and away from the input prompt respectively. In the second column, we perform part segmentation on the scenes using DINOv2. We see that it performs decent on the flower and fortress scenes but poorly on the \trex scene, owing to it being a challenging scene. We see similar results in SAM, where we perform decently well in the first two scenes but miss catching the fine ribs in the \trex scene.
}

\end{figure*}
}

\end{document}